\documentclass{article}
\usepackage[nonatbib,preprint]{neurips_2020}

\usepackage[utf8]{inputenc} 
\usepackage[T1]{fontenc}    
\usepackage{hyperref}       
\usepackage{url}            
\usepackage{booktabs}       
\usepackage{amsfonts}       
\usepackage{nicefrac}       
\usepackage{microtype}      

\usepackage{amsmath,amsfonts,bm}

\def\eqref#1{equation~\ref{#1}}

\def\1{\bm{1}}

\DeclareMathAlphabet{\mathsfit}{\encodingdefault}{\sfdefault}{m}{sl}
\SetMathAlphabet{\mathsfit}{bold}{\encodingdefault}{\sfdefault}{bx}{n}

\DeclareMathOperator*{\argmin}{arg\,min}

\usepackage{url}
\usepackage{amsthm}
\usepackage{amsmath}
\usepackage{ragged2e}
\usepackage{bbm}
\usepackage{mathabx}
\usepackage{float}
\usepackage{subcaption}
\usepackage[labelformat=parens,labelsep=quad,skip=3pt]{caption}
\usepackage{graphicx}

\newtheorem{defi}{Definition}

\newtheorem{lemma}{Lemma}

\let\vec\mathbf

\newcommand{\x}{{\vec x}}
\newcommand{\y}{{\vec y}}

\ifthenelse{\isundefined{\definition}}{\newtheorem{definition}{Definition}}{}
\ifthenelse{\isundefined{\assumption}}{\newtheorem{assumption}{Assumption}}{}
\ifthenelse{\isundefined{\hypothesis}}{\newtheorem{hypothesis}{Hypothesis}}{}
\ifthenelse{\isundefined{\proposition}}{\newtheorem{proposition}{Proposition}}{}
\ifthenelse{\isundefined{\theorem}}{\newtheorem{theorem}{Theorem}}{}
\ifthenelse{\isundefined{\lemma}}{\newtheorem{lemma}{Lemma}}{}
\ifthenelse{\isundefined{\corollary}}{\newtheorem{corollary}{Corollary}}{}
\ifthenelse{\isundefined{\alg}}{\newtheorem{alg}{Algorithm}}{}
\ifthenelse{\isundefined{\example}}{\newtheorem{example}{Example}}{}

\usepackage{nicefrac}

\newcommand{\RomanNumeralCaps}[1]
    {\MakeUppercase{\romannumeral #1}}
\usepackage[toc,page,header]{appendix}
\usepackage{minitoc}
\usepackage{verbatim}

\title{On Data Augmentation and Adversarial Risk:\\An Empirical Analysis}

\author{
  Hamid Eghbal-zadeh$^{1,2}$
  \ \ \ \ \ \ \ \ \ \ \ 
  Khaled Koutini$^2$
  \ \ \ \ \ \ \ \ \ \ \ 
  Paul Primus$^2$
  \ \ \ \ \ \ \ \ \ \ \ \\
  \textbf{Verena Haunschmid}$^2$
  \ \ \ \ \ \ \ \ \ \ \
  \textbf{Michal Lewandowski}$^3$
  \ \ \ \ \ \ \ \ \ \ \
  \textbf{Werner Zellinger}$^3$
  \ \ \ \ \ \ \ \ \ \ \ \\
  \textbf{Bernhard A. Moser}$^3$
  \ \ \ \ \ \ \ \ \ \ \
  \textbf{Gerhard Widmer}$^{1,2}$  \\ \\
{$^1$~LIT Artificial Intelligence Lab}, {Johannes Kepler University, Linz, Austria}\\
{$^2$~Institute of Computational Perception}, {Johannes Kepler University, Linz, Austria}\\
{$^3$~Software Competence Center Hagenberg GmbH, Hagenberg, Austria}\\
  \texttt{hamid.eghbal-zadeh@jku.at}
}

\begin{document}
\doparttoc 
\faketableofcontents 

\maketitle

\begin{abstract}
Data augmentation techniques have become standard practice in deep learning, as it has been shown to greatly improve the generalisation abilities of models.
These techniques rely on different ideas such as invariance-preserving transformations (e.g, expert-defined augmentation), 
statistical heuristics (e.g, Mixup), and learning the data distribution (e.g, GANs).
However, in the adversarial settings it remains unclear under what conditions such data augmentation methods reduce or even worsen the misclassification risk.
In this paper, we therefore analyse the effect of different data augmentation techniques on the adversarial risk by three measures: (a) the well-known risk under adversarial attacks, (b) a new measure of prediction-change stress based on the Laplacian operator, and (c) the influence of training examples on prediction.
The results of our empirical analysis disprove the hypothesis that an improvement in the classification performance induced by a data augmentation is always accompanied by an improvement in the risk under adversarial attack.
Further, our results reveal that the augmented data has more influence than the non-augmented data, on the resulting models. 
Taken together, our results suggest that general-purpose data augmentations that do not take into the account the characteristics of the data and the task, must be applied with care.
\end{abstract}

\section{Introduction}
\label{sec:intro}

Data augmentation is one of the fundamental building blocks of deep learning,
and it has been shown that without it, deep neural networks suffer from various
problems such as lack of generalisation~\cite{perez2017effectiveness,antoniou2018data,yun2019cutmix,berthelot2019mixmatch} and adversarial
vulnerability~\cite{zhang2017mixup,perez2017effectiveness,pmlr-v97-verma19a,adversarianexample_gan_ijcai}.
By affecting a model's behavior outside of the given training data,
any data augmentation strategy introduces a certain \textit{bias} caused by its assumptions about the data, or the task at hand~\cite{battaglia2018relational}.
Some augmentation methods incorporate inductive bias into the model by (classical) invariance-preserving geometric transformations designed by domain experts,
while others rely on statistical heuristics (e.g, Mixup~\cite{zhang2017mixup}) or sampling from a learnt data distribution (e.g, GANs~\cite{antoniou2018data,adversarianexample_gan_ijcai}).

Although it is known that some data augmentation methods reduce classification error and/or adversarial risk, the \emph{relationship} between the error and the adversarial risk, \emph{under different assumptions} imposed by the augmentation strategies has not been extensively studied.
Furthermore, \emph{the level of influence from the augmented data} on the resulting models, in comparison with the influence of the original training data, remains unclear.
These two aspects are particularly important, 
as it has been shown that adversarial examples~\cite{goodfellow2014explaining,kurakin2016adversarial,biggio2018wild,szegedy2013intriguing} --a known issue in deep neural networks-- are the result of features that lie in the data space~\cite{ilyas2019adversarial}, hence the robust or non-robust nature of the features in the data, 
defines the amount of adversarial risk in the resulting models.
Therefore, it is important to study the relationship between the data augmentation and adversarial risk.

In this work, we therefore address two research questions that have been underrepresented:
(a), the qualitative analysis question, what (side) effects do the different data augmentation strategies have on the risk of misclassification of a model compared to the adversarial risk?
and (b), the quantitative analysis question, how to detect and analyse such effects by 
quantitative indicators e.g. by measuring the influence of augmented data versus the original data on the overall model's behavior?

Regarding question (a), extensive experiments with different data augmentation methods,
and with systematic adversarial attacks, reveal that the success of a data augmentation in reducing the classification error of a model, is not an indicator of its capabilities in lowering of the adversarial risk. We observe that while some of the augmentation strategies (such as Mixup and GANs) both improve classification performances (notably to a lower degree), 
one (Mixup) achieves the highest adversarial risk, while the other (GANs) results in models with the lowest risk under attack.
We also observe that expert-introduced augmentations achieve the lowest classification error, and at the same time, they are among the most adversarially robust models.

Regarding question (b), to reveal the underlying reasons behind, we propose a new measure called \textit{prediction-change stress} which provides an estimate on the robustness of model's predictions.
Using this measure, we show that data augmentations that result in models with high prediction-change stress around test examples, 
suffer from high risk under adversarial attacks,
while data augmentation strategies that result in models with low prediction-change stress are the most adversarially robust. Finally, we use `influence functions'~\cite{Koh17} to estimate the amount of influence from original training, and augmented data, for different augmentation methods.
Our influence analysis shows that incorporating data augmentation -- which is a common practice in deep learning--introduces a much higher influence on the final model than the original data itself.
This underpins our final conclusion, that the choice of data augmentation and its assumptions must be adjusted to the characteristics of the data and the task at hand.

Addressing both questions, our empirical analysis is based on three classes of data augmentation in the visual domain, that is \emph{Classical augmentation} via geometric transformations, \emph{Mixup}~\cite{zhang2017mixup}, and \emph{GAN augmentation}, by looking at both classification error and risk under adversarial attacks.

\section{Problem Description}
Throughout this work let $X$ be a random variable on a probability space $(\mathcal{X},\mathcal{A},P)$ with $\mathcal{X}\subset\mathbb{R}^d$, e.g.~images, and denote by $P$ the probability measure of $X$, where $\mathcal{A}$ denotes a sigma algebra on $\mathcal{X}$.
Further, let $l:\mathcal{X}\to\mathcal{Y}$ be a labeling function to a finite set $\mathcal{Y}\subset\mathbb{N}$ of labels, e.g.~$\{1,\ldots,c\}$.

Given a class $\mathcal{F}$ of functions $f:\mathcal{X}\to\mathcal{Y}$ and a sample
\begin{align}
    \label{eq:sample}
    S=\left((\x_1,l(\x_1)),\ldots,(\x_s,l(\x_s))\right)\in (\mathcal{X}\times\mathcal{Y})^s,
\end{align}
of input-label pairs, i.e.~$s$ observations $\x_1,\ldots,\x_s$ independently drawn from $P_X$ with corresponding labels $l(\x_1),\ldots,l(\x_s)$, the problem of \textit{risk minimization} is to find a function $f\in\mathcal{F}$ with low \textit{misclassification risk} $R(f,l):=P(f(X)\neq l(X))$.

One successful method for solving problems of risk minimization is to perform stochastic gradient descent based on some parametric function class $\mathcal{F}$ of neural networks~\cite{lecun2015deep}.
In many practical tasks, this method can be improved by applying so-called \textit{data augmentation} techniques.

In the following, we give a general definition of
data augmentation in a probabilistic setup which permits us to apply an augmentation function to a set of data-label pairs and transform them into some augmented version with a certain probability.

\begin{defi}[Data Augmentation]
\label{def:data_augmentation}
    We call a random function
    \begin{align}
        \label{eq:data_augmentation}
        \begin{split}
            A:(\mathcal{X}\times\mathcal{Y})^s &\to \left\{X\times Y:\mathcal{X}\times\mathcal{Y}\to\mathbb{R}^d\right\}^r
        \end{split}
    \end{align}
    a \textbf{data augmentation}, if it maps a sample $S=((\x_1,l(\x_1)),\ldots,(\x_s,l(\x_s)))\in(\mathcal{X}\times\mathcal{Y})^s$, with measure $P_X$ on $\mathcal{X}$ and labeling function $l:\mathcal{X}\to\mathcal{Y}$, to some vector $A(S)=\left(X_1\times Y_1,\ldots, X_r\times Y_r\right)$ of independent random vectors $X_1\times Y_1,\ldots, X_r\times Y_r:\mathcal{X}\times\mathcal{Y}\to\mathbb{R}^d$ with measure $P_{X_\RomanNumeralCaps{1}\times Y_\RomanNumeralCaps{1}}$ on $\mathcal{X}\times\mathcal{Y}$ and marginal measure $P_{X_\RomanNumeralCaps{1}}$ dominating $P_X$.
\end{defi}

By Definition~\ref{def:data_augmentation}, an \textit{augmented sample} $\tilde S=((\tilde \x_1,\tilde y_1),\ldots,(\tilde \x_s,\tilde y_s))$ can be obtained from a sample $S\in(\mathcal{X}\times\mathcal{Y})^s$ by observing the random variable $A(S)$.
The assumption $P_{X_\RomanNumeralCaps{1}}$ dominating $P_X$ ensures data augmentations take the original sample into account, i.e.~if $P_X(D)>0$ then also $P_{X_\RomanNumeralCaps{1}}(D)>0$ for any measurable $D$.

One example of data augmentation with the goal of enriching the input manifold is the classical (expert-based) data augmentation which combines image cropping, swapping and noise adding.
Another example is GAN-based data augmentation, which aims at generating data from the input distribution~\cite{goodfellow2014generative}.
A third example is Mixup~\cite{zhang2017mixup} data augmentation which aims at generating data in the vicinity of $S$.
See Appendix~\ref{appx:theory_aug_examples} for formalizations based on Definition~\ref{def:data_augmentation}.

It has been noted~\cite{zhang2017mixup,perez2017effectiveness,pmlr-v97-verma19a,adversarianexample_gan_ijcai} that training on augmented samples leads to models with lower risk on \textit{adversarial examples} when compared to models obtained without data augmentation.
Typical measures for \textit{adversarial risk} are the \textit{risk under corrupted inputs}, see e.g.~\cite{mansour2014robust,attias2018improved},
\begin{align}
    \label{eq:corrupted_input_risk}
    R_{\mathrm{cor}}(f,l,\epsilon):=P(\exists\,\x\in B_\epsilon(X) : f(\x)\neq l(X))
\end{align}
with $B_\epsilon(\x):=\{\x'\in\mathbb{R}^d\mid c(\x',\x)\leq\epsilon\}$ for some distance $c:\mathcal{X}\times\mathcal{X}\to\mathbb{R}_+$, the \textit{prediction-change risk},~\cite{SZS+14},
\begin{align}
\label{eq:prediction_change_adv_robustness}
    R_\mathrm{pc}(f,\epsilon):=P(\exists\,\x\in B_\epsilon(X): f(\x)\neq f(X))
\end{align}
and combinations of these concepts~\cite{suggala2018revisiting}.
It is interesting to note, that examples can be constructed where both formulations, Eq.~\ref{eq:corrupted_input_risk} and Eq.~\ref{eq:prediction_change_adv_robustness}, show counter-intuitive behaviour and therefore are better interpreted as approximations of adversarial risk than in terms of  definitions~\cite{DMM18,tsipras2018robustness}.

However, data augmentations as given by Definition~\ref{def:data_augmentation} 
do not necessarily decrease adversarial risk as measured by Eq.~\ref{eq:corrupted_input_risk} and Eq.~\ref{eq:prediction_change_adv_robustness}.
To see this, consider the situation of an optimal classifier $f^* \in \argmin_{f\in\mathcal{F}} R(f,l)$ which is not perfect, i.e.~$f^*\neq l$, due to the limited complexity of the function class $\mathcal{F}$.
Further consider a data augmentation that generates additional data points around some misclassified data from the original data set.
The overweighting of misclassified data might lead to a new classifier $\widetilde{f} \in \mathcal{F}$ that minimizes the loss based on the augmented data while worsening the misclassification risk $R$ and also the adversarial risk as measured by Eq.~\ref{eq:corrupted_input_risk} and Eq.~\ref{eq:prediction_change_adv_robustness}.

The example above considers idealised training scenarios where all augmented samples are taken into account.
However, in most practical cases, not all examples in the training sample have the same influence.
In addition to the measures of Eq.~\ref{eq:corrupted_input_risk} and Eq.~\ref{eq:prediction_change_adv_robustness}, we therefore also look at the \emph{influence} of the augmented samples as an indicator of a change in robustness compared to models obtained without data augmentation.

\subsection{Measuring the Effect of Data Augmentation on Adversarial Risk}
\label{subsec:risk_scores}

For analysing the influence of different data augmentation techniques on the adversarial risk of learned models (available in Section~\ref{sec:empirical_results}), we rely on three measures: (a) the estimated risk under adversarial attacks, (b) a measure for prediction-change stress and (c) a measure for estimating the influence of training examples on models when predicting on unseen data and adversarial examples.
While the former is a common measure for approximating Eq.~\ref{eq:corrupted_input_risk}, the latter two are new in this context.

\subsubsection{Risk under Attack}
\label{subsec:risk_under_attack}

In order to estimate Eq.~\ref{eq:corrupted_input_risk} we rely on corrupted inputs generated by some adversarial attack.
Therefore, we compute a perturbation $\delta$ for an input $\x\in\mathcal{X}$ using an iterative Projected Gradient Descent (PGD) as follows:
\begin{align}
\label{eq:pgd}
    \delta(f,\x) := \mathcal{P}_\epsilon( \delta + \alpha \nabla_\delta L(f(\x+\delta),l(\x)),
\end{align}
where $L:\mathcal{Y}\times \mathcal{Y}\to \mathbb{R}$ is a loss, $\alpha>0$ is the step-size and $\mathcal{P}_\epsilon$ is a projection from the inputs $\mathcal{X}$ into the ball $B_\epsilon(X_0)$, $\epsilon>0$\footnote{We refer to the $\epsilon$ and $\alpha$ used in PGD as $PGD_\epsilon$ and $PGD_\alpha$}.
We then approximate Eq.~\ref{eq:corrupted_input_risk} by computing the misclassification risk on a test sample $\tilde S:=((\x_1',l(\x_1')),\ldots,(\x_s',l(\x_s')))$, i.e.
\begin{align}
    \label{eq:pgd_score}
    \widehat{R}_{\mathrm{cor}}(f,\tilde S,\epsilon):=\frac{1}{s}\sum_{i=1}^s \mathbbm{1}_{f(\x_i'+\delta(f, \x_i'))\neq l(\x_i')},
\end{align}
where $\mathbbm{1}_{a\neq b}=1$ if $a\neq b$ and $\mathbbm{1}_{a\neq b}=0$ otherwise.

\subsubsection{Measure of Prediction-Change Stress}
\label{subsec:pcs_definition}
Due to the phenomenon of concentration of measure \cite{Don00}, 
for high dimension $d$ the Lebesgue measure $\lambda$ in a ball $B^{(d)}_r$ of radius $r$ is concentrated at its surface. That is, for any $\delta>0$, we have 
$\lim_{d \rightarrow \infty} (\lambda( B^{(d)}_r) - \lambda(B^{(d)}_{r-\delta}))/\lambda(B^{(d)}_r) = 1$ given any $l_p$-norm ($p\in [1, \infty]$) we take.
As the dimension $d$ in most applications is high, the concentration of measure effect motivates to approximate  Eq.~\ref{eq:corrupted_input_risk} and 
Eq.~\ref{eq:prediction_change_adv_robustness} by sampling only from the shell 
$\partial B_{\epsilon}$ (points at distance $\epsilon$ from the center of the ball) instead of sampling from the full ball $B_{\epsilon}$. 

In order to estimate Eq.~\ref{eq:prediction_change_adv_robustness}, we consider only the points on the boundary $\partial B_{\epsilon}$ of the ball $B_{\epsilon}$ and introduce the {\it measure of prediction-change stress} on a sample $\tilde S:=((\x_1',l(\x_1')),\ldots,(\x_s',l(\x_s')))$ by \begin{align}
\label{eq:stress}
\widehat{\mbox{stress}}_{\mathrm{pc}}(f,\tilde S,\epsilon):= \frac{1}{s}\sum_{i=1}^s \frac{1}{n}\sum_{j=1}^n \mathbbm{1}_{f(\x_i')\neq f(\y_{ij})}
\end{align}
where $\y_{i1},\ldots,\y_{in}$ are uniformly sampled from $\partial B_{\epsilon}(\x_i')$. Note that 
$\widehat{\mbox{stress}}_{\mathrm{pc}}(f,\tilde S,\epsilon) \approx 
R_{\mathrm{pc}}(f,\epsilon)$ for sufficiently high dimension $d$ and sample size $s$.

Eq.~\ref{eq:stress} allows the interpretation of stress in analogy to physics where mechanical stress is a measure of an external force acting over the cross sectional area of an object causing deformation of its geometric shape. 
This interpretation is closely related to the Laplacian operator of a function $f$ at a point $\x_0$, which can be understood as limit of means over the differences of $f(\x_0)$ and the boundary values $f(\x) \in \partial B_{\epsilon}(\x_0)$ of its surrounding $\epsilon$-balls~\cite{Sty2015}. 
Note that Eq.~\ref{eq:stress} can be understood as estimation of the Laplacian of the indicator function w.r.t.~the center prediction $f(\x)$. Utilizing this physical interpretation therefore justifies interpreting adversarial points as imposing stress on the shape of the decision surface. The more {\it prediction-change stress} the less robust is the model at point $\x$.

\subsection{Influence of Augmented Samples}
\label{subsec:influence_score}

To analyse the effect of augmented samples on the prediction of some model $f$, we rely on the influence score introduced in~\cite{Koh17} for both original and augmented training data.
The influence $\widehat{I}(\x, \x_\text{test})$ of an input $(\x,l(\x))\in S$ in the training sample $S$ on a new test example $\x_\text{test}\not\in{S}$ given a model $f_\theta$ with parameters $\theta$ is defined by:
\begin{align}
\label{eq:influence}
\widehat{I}(\x, \x_\text{test}) :=
-\nabla_\theta L(f_{\hat\theta}( \x_\text{test}),l(\x_\text{test})) ^\top H_{\hat\theta}^{-1} \nabla_\theta L(f_{\hat\theta}(\x),l(\x)),
\end{align}
where $H$ is the Hessian, $L:\mathcal{Y}\times \mathcal{Y}\to\mathbb{R}$ is a loss, and $\hat\theta$ is the minimizer of the empirical risk $\frac{1}{s}\sum_{i=1}^s  L(f_{\theta}(\x_i),l(\x_i))$.
The influence score as calculated by Eq.~\ref{eq:influence} can be interpreted as measuring how much the loss computed on a test example changes if a training sample is up-weighted.

In our empirical evaluations, we analyse the distribution of the influence values $\widehat{I}(\x, \x_\text{test})$ corresponding to different training examples $\x$ and test examples $\x_\text{test}$,  as measured by Eq.~\ref{eq:influence} for different data augmentation techniques.
Additionally, we analyse the distributions of influence values $\widehat{I}(\x, \x_\text{adv. test})$, 
for training examples $\x$ given the adversarial test examples $\x_\text{adv. test}$.

\section{Empirical Evaluations}
\label{sec:empirical_results}

\subsection{Goals and Questions}
\label{subsec:exp_goals_questions}

The main body of the experiments is built upon the task of image classification, as it is a widely-used and established task in machine learning.
We train Resnet50~\cite{he2016deep} image classification models on CIFAR10~\cite{krizhevsky2009learning}.
In each experiment, we control the amount of augmentation applied during the training.
More specifically, we apply a certain data augmentation technique, with a certain probability, during the training of a model.
For each trained model, we then report the test classification error, risk under adversarial attack, prediction-change stress, and influence scores.

Our goal is to understand the effect of each data augmentation on the resulting models as the probability of that data augmentation increases, 
and discover the characteristics of the trained models, with respect to the risk under adversarial attack, and the classification performance.
As the probability of a data augmentation varies between 0 and 1, we can analyse its relationship with a risk, which enables us to study the correlation between the two; and discover the bigger picture.

Given the experimental framework explained above, we are trying to address the following questions:
1) Which data augmentations lower the risk under adversarial attack, 
while reducing the classification error?
2) Is an improvement in the classification performance induced by a data augmentation, always accompanied by an improvement in the risk under adversarial attack?
3) Is the proposed prediction-change stress, a good indicator for the risk under adversarial attacks?
and 4) Does augmented data have a higher \emph{influence} on the final models, 
in comparison with the non-augmented data?
In the upcoming sections, we provide the results of our experiments, and answer the questions above. 
Additional discussions on our empirical results, along with extended experiments are provided in Section~\ref{appx:extended_empirical_results} of the Appendix.

\subsection{Experimental Setup}
\label{subsec:exp_setup}

\textbf{Applying data augmentation}:
In our empirical analysis, we investigate 3 data augmentations, namely, Classical, Mixup, and GAN augmentation.
Each data augmentation is applied independently, to measure its impact on the models.
During training of each model, a data augmentation, with a certain probability is selected. 
The augmentation is applied on each training example with the given probability.
This means that if the probability of augmentation is set to $0.5$, each training sample has $50\%$ chance of being augmented with the given augmentation method.
The probability of data augmentation for each augmentation method, is varied between 0 and 1 (with $0.1$ increments), and is referred to as $P_{Aug}$.

\textbf{Classification experiments}:
For each data augmentation, and for a given augmentation probability, a Resnet50 model is trained for 200 epochs using SGD, weight-decay, and learning-rate schedule.
Each experiment is repeated three times, where in each experiment the models were initialised using a different random seed.
The details of the training procedure is provided in the Appendix Section~\ref{subsec:training_setup}.

\textbf{Adversarial attacks}:
After a model is fully trained, four white-box PGD attacks are performed on the test set with $\text{PGD}_{\epsilon}$ of $0.5$ and $0.25$, and with $10$ and $100$ iterations,
and $\text{PGD}_\alpha$ is always set to $\nicefrac{\text{PGD}_\epsilon}{5}$.
$\text{PGD}_\epsilon$ is the radius for threat model, while $\text{PGD}_\alpha$ is the step size for adversarial attacks.
More details about the attacks are given in the Appendix Section~\ref{appx:attack_setup}.

\textbf{Evaluation measures}:
The trained models are evaluated with the classification error on the test set, 
\emph{Risk Under Adversarial Attack} as defined in Section~\ref{subsec:risk_under_attack}, \emph{Prediction-change Stress} as detailed in Section~\ref{subsec:pcs_definition}, 
and \emph{Influence Scores} as explained in Section~\ref{subsec:influence_score}.

In our empirical analysis using the prediction-change stress, we report the results using $\partial B_{\varepsilon} = \{x\in \mathbb{R}^d\mid \sum_{k=1}^d x_k^2 = \varepsilon^2\}$ ($l_2$ norm), 
and according to Eq.~\ref{eq:stress}.
We provide the prediction-change stress for the 50k non-augmented, and 50k augmented examples randomly chosen from the training set, and for all the 10k test examples.
We randomly sample 1K points from the sphere surface $\partial B_{\varepsilon}(\x)$ around each data point $\x$.
We evaluate models trained with all augmentation methods, for $P_{Aug}\in \{0.1, 0.5, 0.9\}$.

In the influence analysis experiments, we compute the influence scores of 50k non-augmented, and 50k augmented training examples on 200 randomly chosen test examples and their adversarially-perturbed counterparts. 
We calculate the influence scores for models trained with Classical-, MixUp-, and GAN-augmentations and $P_{Aug}\in \{0.1, 0.5, 0.9\}$. 
For the creation of the adversarially perturbed test sets, we start with a fixed set of randomly selected 20 test samples per class, and conduct a PGD attack with 10 iterations using $\text{PGD}_\epsilon=0.25$ and $\text{PGD}_\alpha=0.05$.

\textbf{Classical Augmentation}:
We apply horizontal flipping, changing the brightness, contrast and saturation, random cropping, and random rotation, which are known expert-introduced data augmentations for the CIFAR10 dataset.

\textbf{GAN Augmentation}:
We train label-conditional GANs~\cite{mirza2014conditional}.
Each generator is conditioned on the one-hot vector of the label, which is concatenated to the noise vector,
and is trained to generate samples that match the conditioned class label.
We test two different GAN objectives, namely, non-saturating GAN~\cite{goodfellow2014generative},
and Wasserstein GAN with gradient penalty~\cite{gulrajani2017improved}.
Both models are trained to convergence, and both achieved state-of-the-art  Fr\'{e}chet Inception Distance (FID)~\cite{heusel2017gans}.
Randomly generated samples are shown in Figure~\ref{fig:gan_samples} in the Appendix~\ref{appx:gan_setup}.
More details about the attacks are given in the Appendix Section~\ref{appx:gan_setup}.

\textbf{Mixup Augmentation}:
We apply Mixup with $\text{Mixup}_\alpha=1$, as it was used for CIFAR10 experiments in~\cite{zhang2017mixup}, where $\text{Mixup}_\alpha$ is the hyperparameter that controls the beta distribution.

\subsection{Results for the Classification Error and Risk Under Attack}
\label{subsec:clferr_rua}

\begin{figure}[h]
\centering
\includegraphics[width=\linewidth]{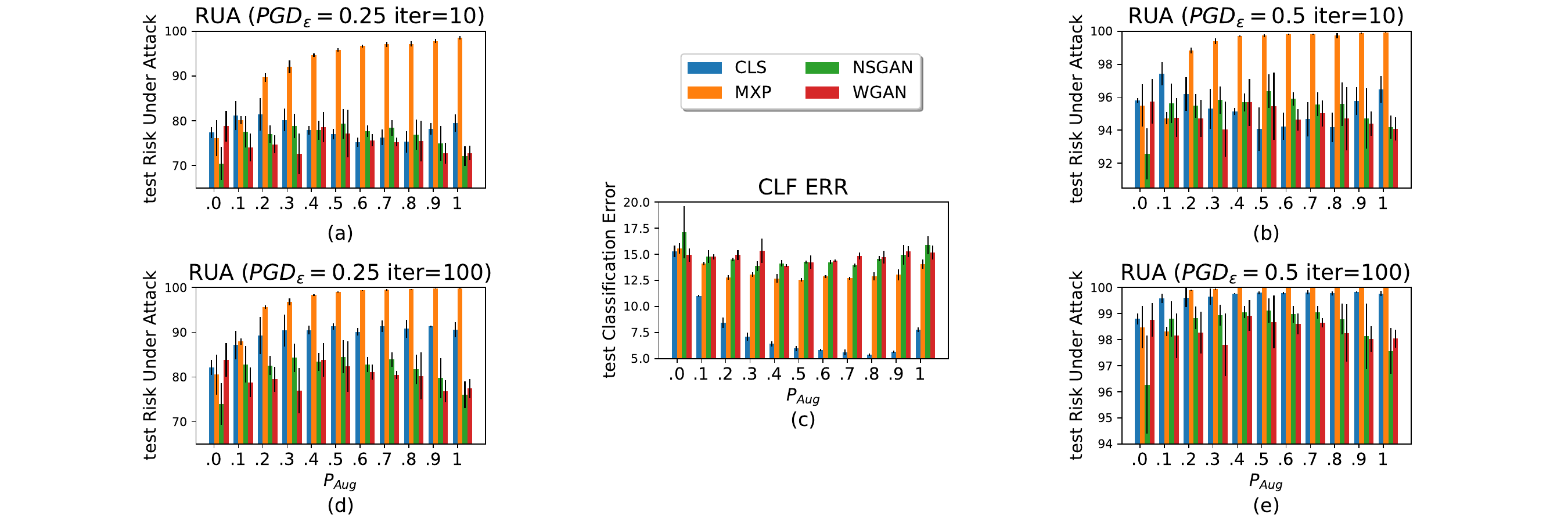}
\caption{
Comparison between the classification error and Risk Under Adversarial Attack (RUA) of different augmentation methods on the test set of CIFAR10.
}
\label{fig:clf_results}
\end{figure}

In this section, we report the classification error, and risk under several adversarial attacks, for different data augmentations, and different augmentation probabilities.
As can be seen in Figure~\ref{fig:clf_results}.c, Classical data augmentation achieves the lowest classification error, and surpasses Mixup and GAN augmentations by a large margin.
Our results show that Mixup can lower the classification error, 
while both GANs do not have a significant effect on the classification error.
We can observe that in all methods, the error declines to a certain degree, as the probability of augmentation increases, then increases again (although for GAN-augmentations, this is less visible compared to the others).
These results suggests that Classical data augmentation is the most successful in lowering the classification error, 
while GAN augmentation has almost no effect in the classification error. 
Mixup shows some success in lowering the classification error, however it achieves a higher error compared to Classical augmentation.

In contrast to the classification error, we can see in Figures~\ref{fig:clf_results}.a,~\ref{fig:clf_results}.b,~\ref{fig:clf_results}.d,~\ref{fig:clf_results}.e that applying Mixup significantly increases the risk under adversarial attacks, in all the 4 PGD attacks.
We can additionally observe that the GAN-augmentations, and the Classical augmentation achieve significantly lower risk under adversarial attacks in comparison with Mixup.
As the strength of the attack increases (larger $\text{PGD}_{\epsilon}$, more iterations), the gap between different augmentation closes, as all models are vulnerable to adversarial attacks.
The difference beween augmentations can be better observed in weaker attacks (10 iterations).
In Figures~\ref{fig:clf_results}.a,~\ref{fig:clf_results}.b we can observe a large gap between the risk under attack in Mixup, and other augmentation methods (Classical and GAN-augmentations).

As discussed, Mixup and Classical augmentation reduce the classification error, while GAN-augmentations do not have a significant effect on the classification performance.
In contrast, Mixup augmentation results in a significant increase in the risk under adversarial attack, while Classical and GAN augmentations incorporate adversarial robustness to the models to some extent.
It can be seen that in attacks with 10 iterations (Figures~\ref{fig:clf_results}.a,~\ref{fig:clf_results}.b) increasing the augmentation probability does not result in an increase in risk under adversarial attacks for Classical and GAN-augmentations.
In stronger attacks (100 iterations shown in Figures~\ref{fig:clf_results}.d,~\ref{fig:clf_results}.e), we can observe an increase in the risk under attack with the augmentation probability for both Classical and Mixup, although Mixup achieves a significantly higher risk.

We also observe that while for some augmentations the risk under adversarial attack rises when the classification error declines (Figures~\ref{fig:clf_results}.a (Mixup), \ref{fig:clf_results}.b (Classic, Mixup), \ref{fig:clf_results}.d (Classic, Mixup))\footnote{This phenomenon is known as the \emph{trade-off between the classification error and adversarial risk}~\cite{tsipras2018robustness}},
in Figure~\ref{fig:clf_results}.e, this is not the case; which can be due to the strength of the attack.
All in all, our results disprove the hypothesis that an improvement in the classification performance induced by a data augmentation, is always accompanied by an improvement in the risk under adversarial attack (a counter example is Mixup).

\subsection{Results for the Prediction-Change Stress}
\label{subsec:pcs}

\begin{figure}[h]
	\centering
	\includegraphics[width=1.03\textwidth]{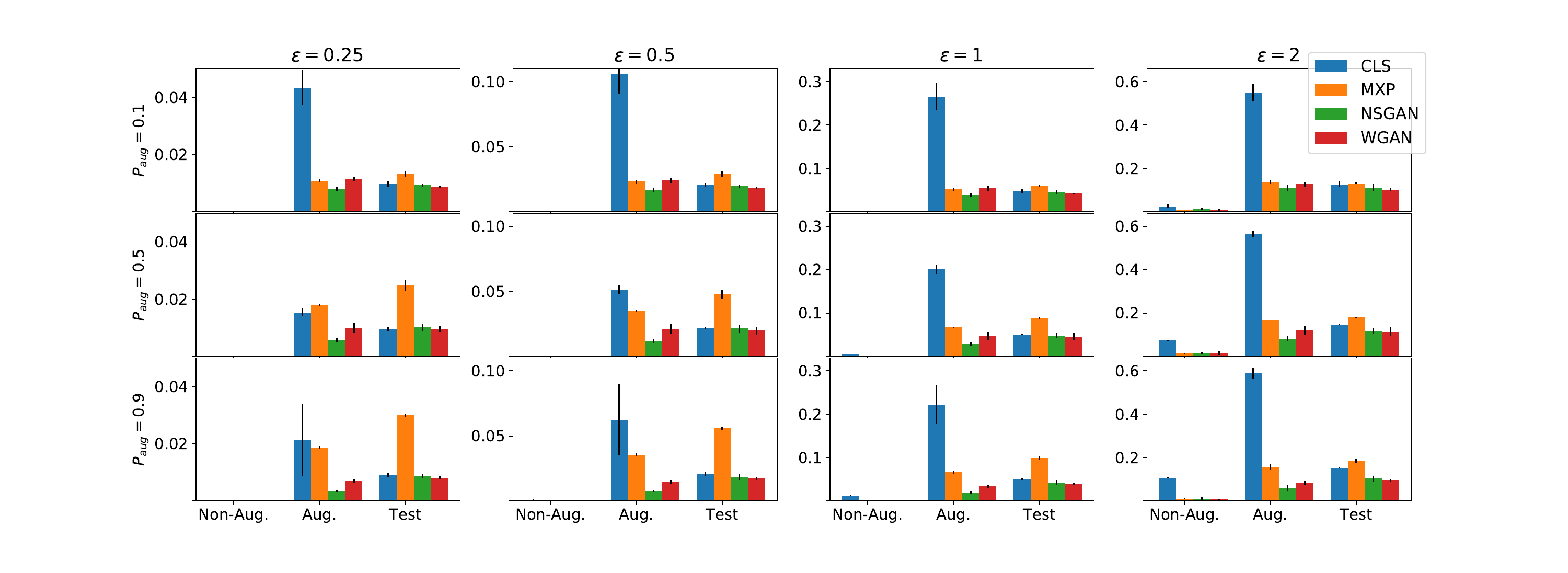}
	\caption{
		Comparison between the Prediction-Change Stress of models with different augmentations. 
		The y-axis indicates the different sets.
		The non-Aug and Aug. refer to the training set.
	}
	\label{fig:avg_stress_results}
\end{figure}

Figure~\ref{fig:avg_stress_results} shows 
the prediction-change stress over the original non-augmented training, augmented-training and the testing datasets for $\epsilon \in \{0.25, 0.5, 1, 2\}$. 

\textbf{Non-augmented training data}:
As can be seen, the results show that the prediction-change stress around the non-augmented training data is significantly smaller, compared to both the augmented training, and the test data.
These results suggest that the trained models have more consistent predictions in the neighborhood around the non-augmented training data.

\textbf{Augmented training data}:
For the augmented data, we can see that in models with GAN and Classical augmentations the prediction-change stress follows a decreasing trend as $P_{Aug}$ increases (especially, from 0.1 to 0.5).
This trend is the opposite for models trained with Mixup, where the prediction-change stress is increased with $P_{Aug}$.
This suggests that Classical and GAN-augmentations makes the model predictions more consistent around the augmented samples, while Mixup fails to achieve this.

\textbf{Test data}:
On the test set, we can see that the difference between different data augmentations are minimal, when the augmentation probability is low ($0.1$). 
As $P_{Aug}$ increases, we can observe that the Mixup augmentation results in models with a larger prediction-change stress around the testing points, compared to the other augmentations.
These result shows that models trained with Mixup have a more inconsistent predictions around the test examples.
Therefore, it is easier to find an example in a small neighborhood of test examples, such that the predictions of such models are changed.
These results are also aligned with the results presented in Figure~\ref{fig:clf_results}, as models trained with Mixup achieve the highest risk under adversarial attacks.
Based on this result, we consider the prediction-change stress as an indicator of the risk under adversarial attacks, hence, an evaluation measure of models in the context of adversarial robustness.

Comparing different values for $\epsilon$, we observe that overall, applying Mixup results in higher prediction-change stress around test examples, compared to other augmentations.
We can also observe a trend that while for smaller  $\epsilon$ values, the difference in the prediction-change stress is higher in models augmented with Mixup, as the $\epsilon$ increases, the gap between the prediction-change stress closes.
This observation is also on par with the results reported in Figure~\ref{fig:clf_results}, as the difference between risk under adversarial attacks in models augmented using different approaches reduced by increasing the strength of the attack.
Additional results on the prediction-change stress are provided in Section~\ref{appx:extended_prediction_change_stress_results} in the Appendix.

\subsection{Results for the Influence Scores}
\label{subsec:is}

\begin{figure}[h]
    \begin{subfigure}[1a]{.24\linewidth}
          \frame{\includegraphics[width=.8\linewidth]{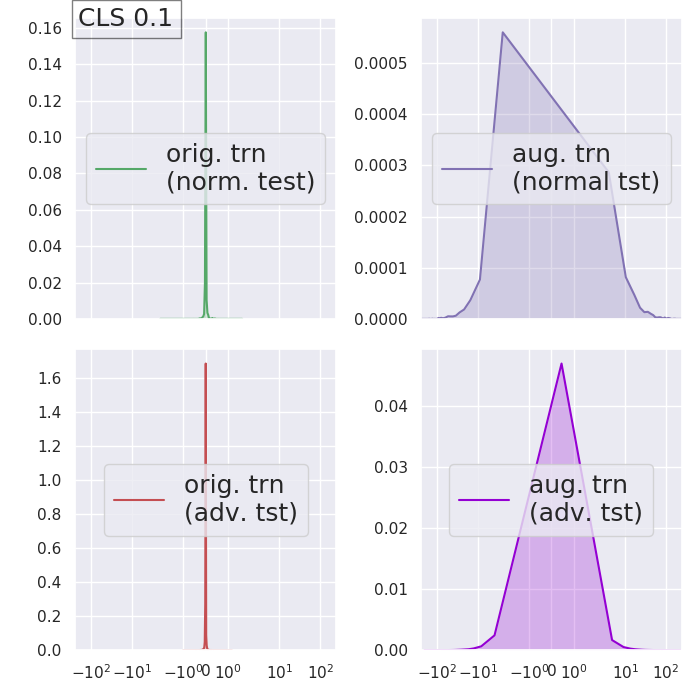}}
    \end{subfigure}
    \hfil
    \begin{subfigure}[2a]{.24\linewidth}
          \frame{\includegraphics[width=.8\linewidth]{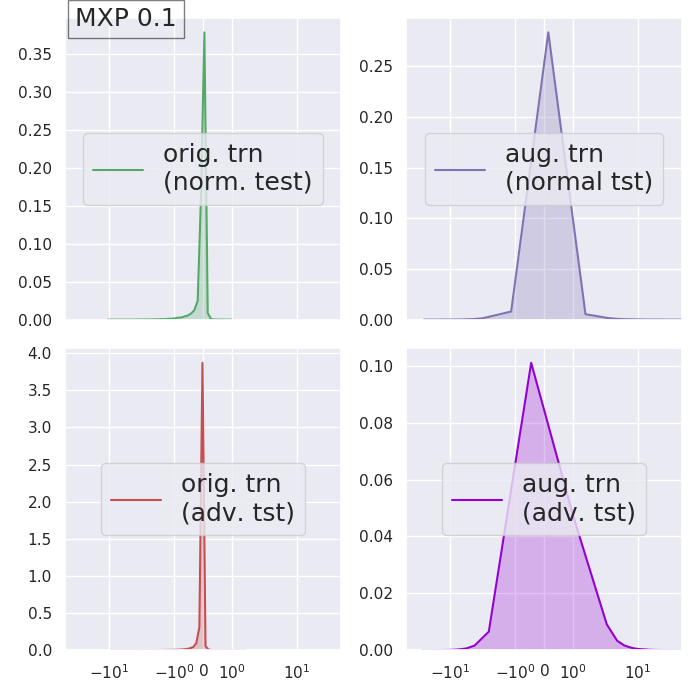}}
    \end{subfigure}
    \hfil
    \begin{subfigure}[3a]{.24\linewidth}
          \frame{\includegraphics[width=.8\linewidth]{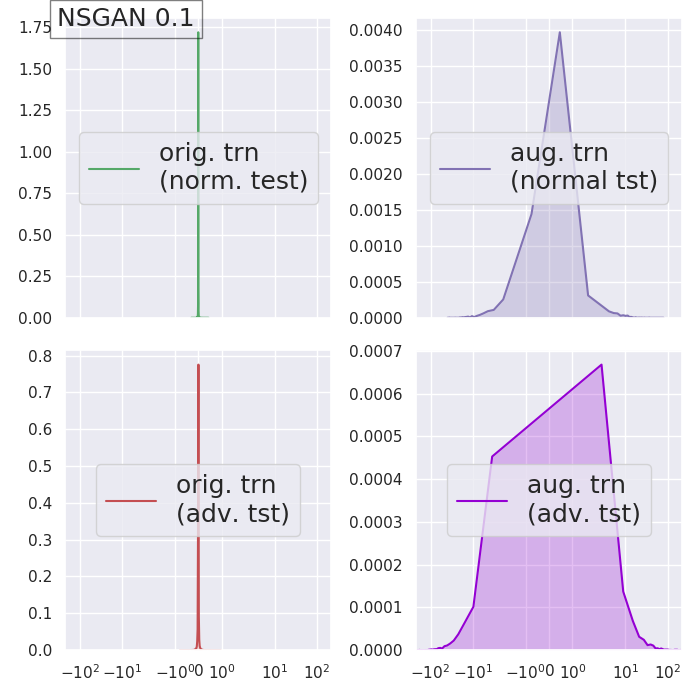}}
    \end{subfigure} 
    \hfil
    \begin{subfigure}[4a]{.24\linewidth}
          \frame{\includegraphics[width=.8\linewidth]{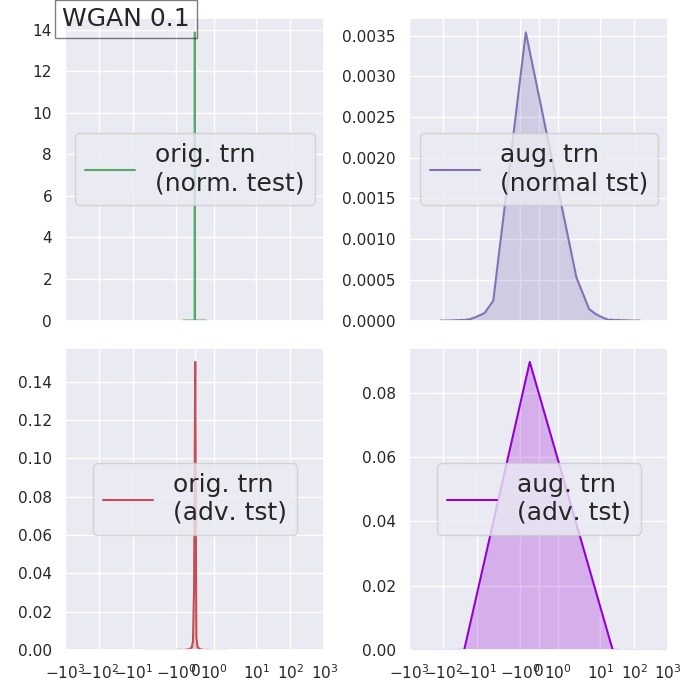}}
    \end{subfigure}
    \vspace{1mm}

    
    \begin{subfigure}[1c]{.24\linewidth}
        
          \frame{\includegraphics[width=.8\linewidth]{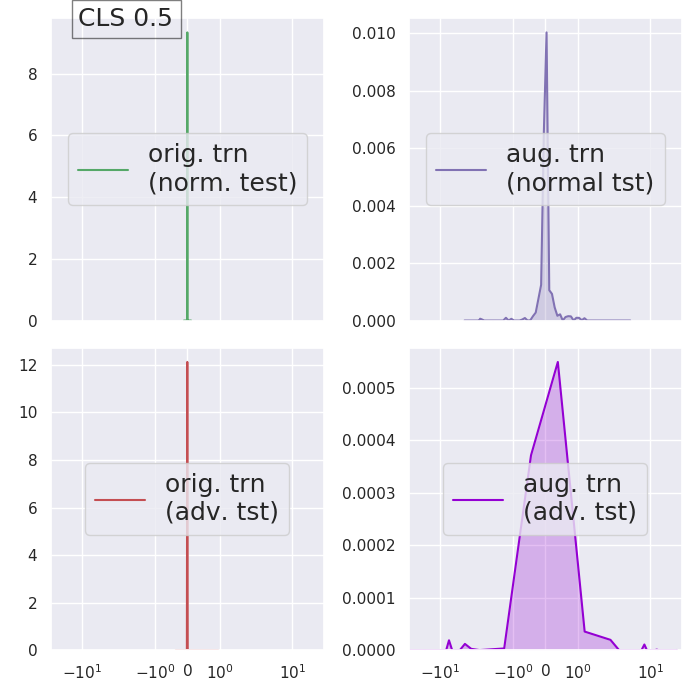}}

    \end{subfigure}
    \hfil
    \begin{subfigure}[2c]{.24\linewidth}
        
          \frame{\includegraphics[width=.8\linewidth]{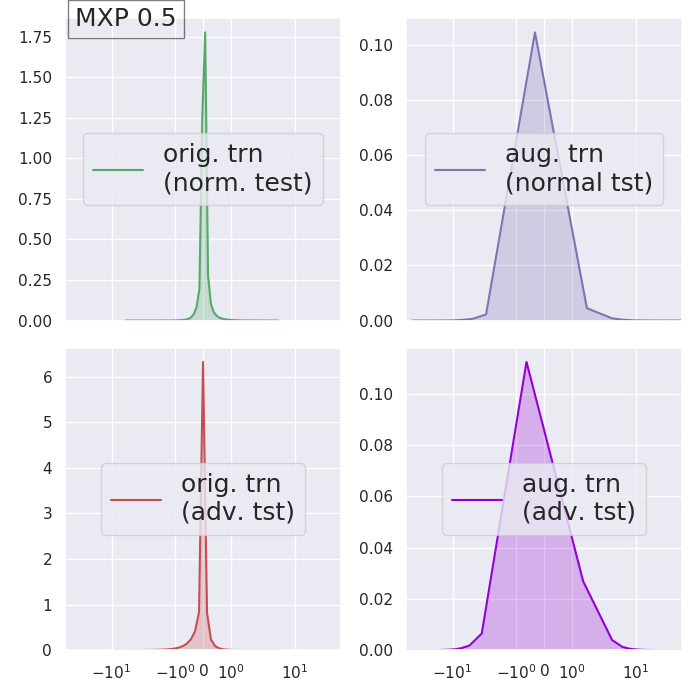}}

    \end{subfigure}
    \hfil
    \begin{subfigure}[3c]{.24\linewidth}
          \frame{\includegraphics[width=.8\linewidth]{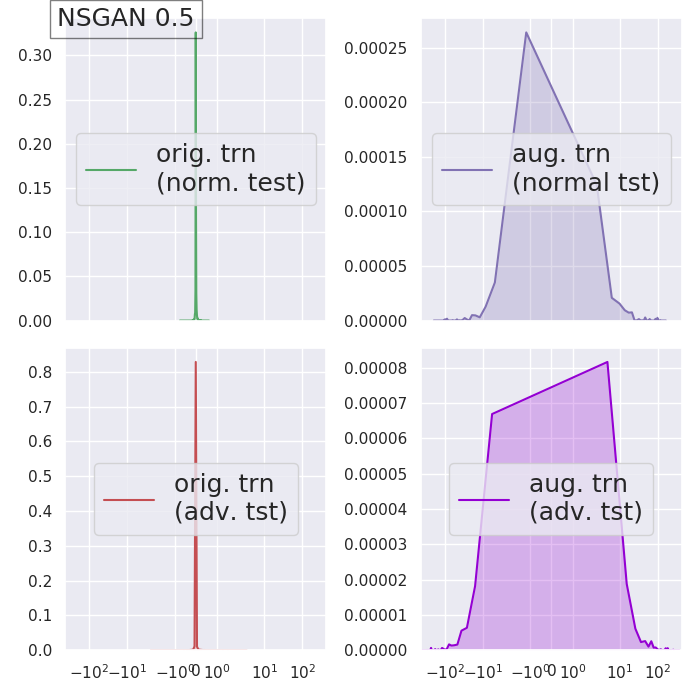}}
    \end{subfigure}
    \hfil
    \begin{subfigure}[4c]{.24\linewidth}
          \frame{\includegraphics[width=.8\linewidth]{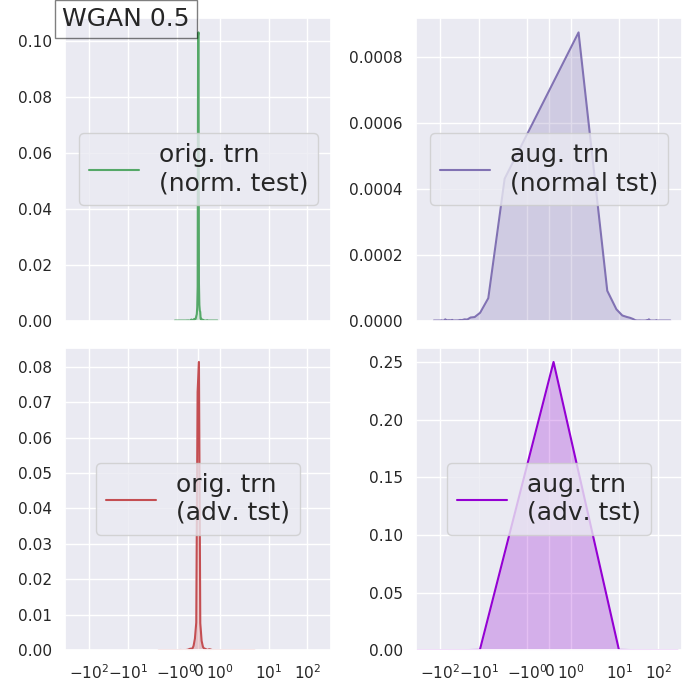}}
    \end{subfigure}
    \vspace{1mm}
    
    \begin{subfigure}[1e]{.24\linewidth}
          \frame{\includegraphics[width=.8\linewidth]{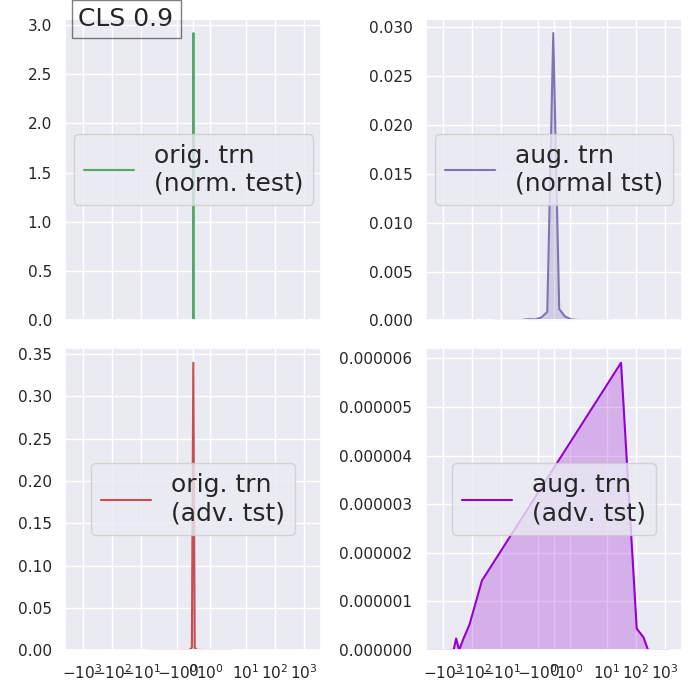}}
    \end{subfigure}
    \hfil
    \begin{subfigure}[2e]{.24\linewidth}
          \frame{\includegraphics[width=.8\linewidth]{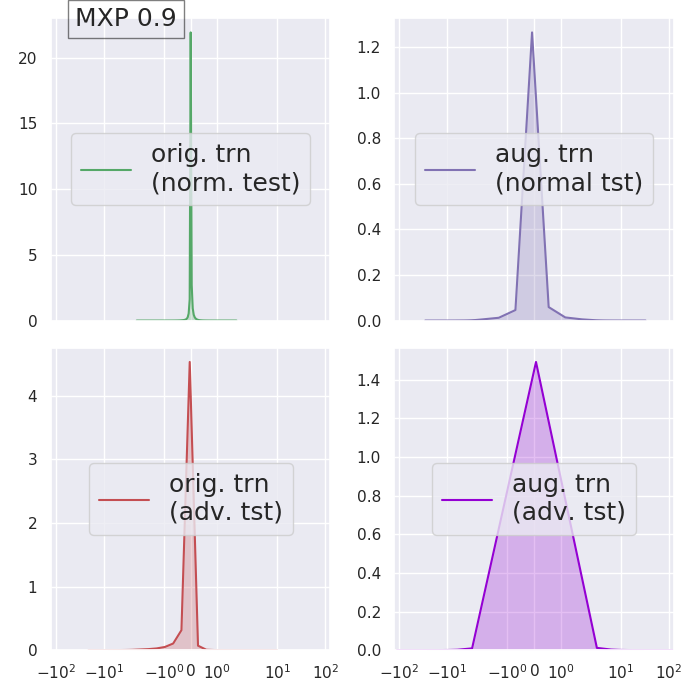}}
    \end{subfigure}
    \hfil
    \begin{subfigure}[3e]{.24\linewidth}
         \frame{ \includegraphics[width=.8\linewidth]{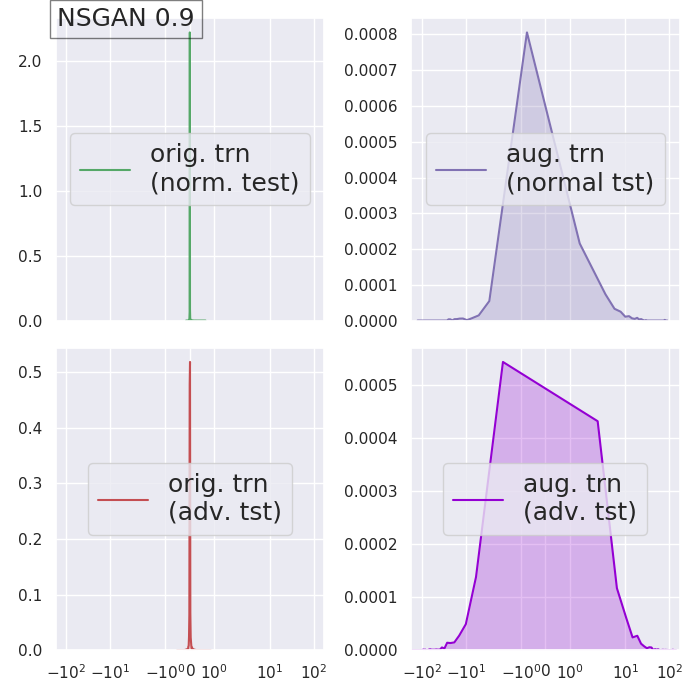}}
    \end{subfigure}
    \hfil
    \begin{subfigure}[4e]{.24\linewidth}
          \frame{\includegraphics[width=.8\linewidth]{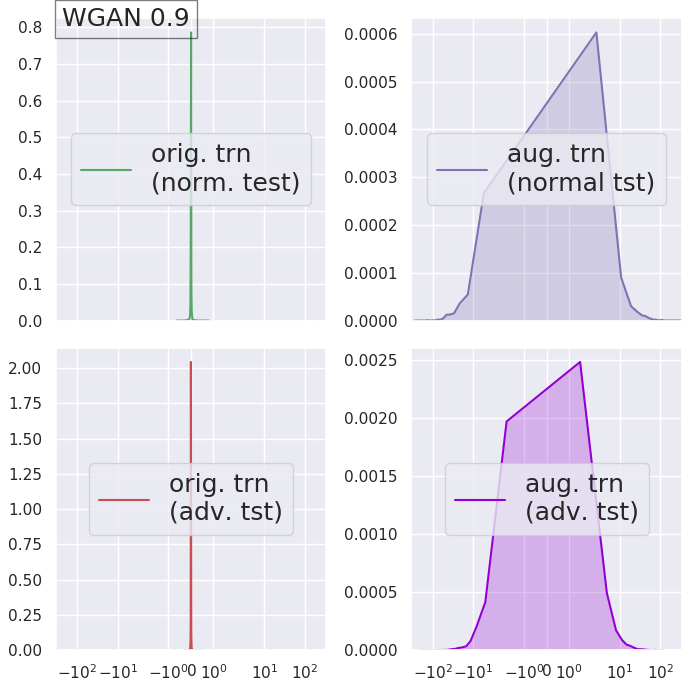}}
    \end{subfigure}
\caption{Comparison between the distribution of original and augmented influence values from the training data, when predicting on the real and the adversarial test set of CIFAR10.
}
\label{fig:influence_dist_results}
\end{figure}

Figure~\ref{fig:influence_dist_results} compares the distribution of influence scores for non-augmented and augmented training samples, given test samples and their adversarially perturbed counterparts. What stands out in this figure is that, across all augmentation techniques and percentages, influence scores of augmented training samples are spread over a broader range compared to non-augmented training samples. These results suggest that, in general, augmented training data contribute stronger towards increasing and decreasing the loss on both original and adversarially perturbed training samples; hence having more influence on training of the models. 

This difference in range can be explained by the incorporated randomness in the augmented data, as opposed to the non-augmented data.
Since the non-augmented training data is deterministic, they have been `seen' by the model multiple times during training.
Consequently, the the average gradient-norm and the loss for the non-augmented examples is overall lower compared to the (previously unseen) augmented examples (see Section~\ref{appx:extended_influence_results} in the Appendix); considering that each augmented data is different from the others, as they are sampled from a stochastic function.
This explanation is consistent with our Prediction-Change Stress results (Section \ref{subsec:pcs}), where we observe significantly lower prediction change stress on the non-augmented training samples.

Our findings suggest that augmented data are more influential during training in comparison to non-augmented data, and therefor data augmentation must always be applied by carefully taking the data and the task characteristics into consideration.
The extended influence results are provided in Section~\ref{appx:extended_influence_results} in the Appendix.

\section{Conclusion}
\label{sec:conclusion}
In this paper, we have analysed the effect of data augmentation techniques on the adversarial risks, as well as their influence on the final models.
Our empirical analysis showed that while some data augmentations can reduce classification error,
this improvement in the classification performance is not always accompanied by an improvement in the risk under adversarial attack.
We proposed a new measure, namely the Prediction-change Stress, that can be used to analyse the behaviour of models in an adversarial setting, and we showed that it strongly correlates with risk under adversarial attack.
Finally, by incorporating influence functions we showed that regardless of their effects on the classification performance and risk under adversarial attack, augmented data has a significantly higher influence on the models compared to non-augmented data.
Hence, the choice of data augmentation and its assumptions must be adjusted to the characteristics of the data and the task at hand, 
to minimise the adversarial risk of the resulting models.

\clearpage\null
\section*{Broader Impact}
This study is intended to raise awareness of possible undesirable side effects for some of the widely used techniques in deep learning, as well as providing case studies for the follow-up basic research in the direction of robustness, analysis and understanding of deep neural networks.

The contributions offered by our paper are not new methods or algorithms, but new insights into existing methods and their vulnerabilities, along with the definition of new measures that help us gain such insights.
We view this as a (small) step towards more reliable and trustworthy
machine learning algorithms. In a broader context, a deeper understanding of a (by now) pervasive technology such as deep learning may not change anything about its biases or misuse potential in real-world applications, but it does hold the promise of making such models and their decision basis more transparent.

\section*{Acknowledgments}
This work has been supported by the COMET-K2 Center of the Linz Center of Mechatronics (LCM) funded by the Austrian federal
government and the federal state of Upper Austria, and has been partly funded by BMK, BMDW, and the Province of Upper Austria in the frame of the COMET Programme managed by FFG in the COMET Module S3AI.

\medskip

\small
\bibliography{refs}
\bibliographystyle{plain}

\newpage
\appendix
\addcontentsline{toc}{section}{Appendix} 
\part{} 
\parttoc 

\section{Extended Results and Setup}
\label{appendix}
In this section, we present our extended results, including:
\begin{enumerate}
    \item Risk under adversarial attack with $l_{inf}$ norm (see Section~\ref{appx:extended_empirical_results}).
    \item Prediction-change stress using $\partial B_{\varepsilon}(\x)$ with $l_{inf}$ norm (see Section~\ref{appx:extended_prediction_change_stress_results}).
    \item The distribution of the average gradient-norm on the augmented and non-augmented training examples (see Section~\ref{appx:extended_influence_results}).
    \item The distribution of the loss for augmented and non-augmented training examples (see Section~\ref{appx:extended_influence_results}).
    \item Examples of various augmentation methods based on our definition of augmentation (see Section~\ref{appx:theory_aug_examples}).
\end{enumerate}

The remainder of this section is as follows. Section~\ref{appx:theory_aug_examples} provides examples of different augmentation methods based on Definition~\ref{def:data_augmentation}.
Section~\ref{subsec:training_setup} describes the setup used in the training of the models, and provides more details on various aspects such as training image classification models, and applying data augmentation.
The details of the adversarial attacks are provided in Section~\ref{appx:attack_setup}.
More information about our GAN training are given in Section~\ref{appx:gan_setup}. 
Our extended empirical results can be found in Sections~\ref{appx:extended_empirical_results},~\ref{appx:extended_influence_results}, and \ref{appx:extended_prediction_change_stress_results}.
The network architectures used in the GAN training are shown in Section~\ref{appx:network_architectures}, and finally the libraries and tools used in this study are detailed in Section~\ref{appx:sec:tools_libs}.

\subsection{Augmentation Examples}
\label{appx:theory_aug_examples}
\textbf{Noise adding} can be formalized as data augmentation $A:(\mathcal{X}\times\mathcal{Y})^s \to \left\{\mathcal{X}\times\mathcal{Y}\to\mathbb{R}^d\right\}^r$ by
    \begin{equation}
        A(S)=\left((\x_1 + Z_1, l(\x_1)),\ldots,(\x_r + Z_r, l(\x_r)) \right),
        \label{eq:noise_adding}
    \end{equation}
    where $Z_1,\ldots,Z_r$ are i.i.d random variables on $[0,\epsilon]$.

\textbf{Random cropping} parts of some vector $\x\in\mathcal{X}$ can be realized by zeroing a random number of the ``outer'' elements by means of

    \begin{align}
    \label{eq:cropping}
         A(S)=\left(  \big( \mathbbm{1}_{(U_{11},U_{21})}(\x_1) , l(\x_1)\big),\ldots,\big( \mathbbm{1}_{(U_{1r},U_{2r})}(\x_r), l(\x_r)\big) \right),
    \end{align}
    where ${\mathbbm{1}_{(U,U')}(\x)=(0,\ldots,0,x_{U},\ldots,x_{d-U'},0,\ldots,0)}$ and $U_{11},\ldots, U_{2r}$ are random variables on $\{1,\ldots,r\}$.

     \textbf{Permuting}, e.g.~swapping, the elements of a vector $\x=(x_1,\ldots,x_d)\in\mathcal{X}$ can also be formalized as augmentation, e.g.~by
    \begin{align}
    \label{eq:cropping}
    A(S) = \left(\left((x_{\pi_1(1)},\ldots,x_{\pi_1(d)}), l(\x_1)\right),\ldots,\left((x_{\pi_r(1)},\ldots,x_{\pi_r(d)}), l(\x_r)\right) \right)
    \end{align}
    where $\pi_1,\ldots,\pi_r:\{1,\ldots,d\}\to\{1,\ldots,d\}$ are permutation functions.
    
The following Lemma allows to express data augmentations as composition of data augmentations.
\begin{lemma}
\label{lem:composition}
If $A$ and $B$ are  data augmentations then $A\circ B$ is also a data augmentation.
\end{lemma}

\begin{proof}
\label{proof:composition}
If $A$ and $B$ are both data augmentations then $A:(\mathcal{X}\times\mathcal{Y})^{s_1} \to \left\{\mathcal{X}\times\mathcal{Y}\to\mathbb{R}^{d}\right\}^{r_1}$ and $B:(\mathcal{X}\times\mathcal{Y})^{s_2} \to \left\{\mathcal{X}\times\mathcal{Y}\to\mathbb{R}^{d}\right\}^{r_2}$ for some $s_1,s_2,r_1,r_2\in\mathbb{N}$.
Then the composition $A\circ B$ is such that
\begin{equation*}
    A\circ B:\left(\mathcal{X}\times\mathcal{Y}\right)^{s_2}\to \left\{\mathcal{X}\times\mathcal{Y}\to\mathbb{R}^{d}\right\}^{r_1}.
\end{equation*}
It remains to show that the marginal measure $P_{X_\RomanNumeralCaps{1}'}^{r_1}$ of some vector $A(\tilde S)$, resulting from a sample $\tilde S$ with measure $P_{X_\RomanNumeralCaps{1}}^{s_1}$ augmented by $A$, dominates the measure $P_X^{s_2}$ of some original sample $S$. By assumption, ${P_{X_\RomanNumeralCaps{1}^{'}}^{r_1}(D)=0}$ implies  $P_{X_\RomanNumeralCaps{1}}^{s_1}(D)=0$ and $P_{X_\RomanNumeralCaps{1}}^{r_2}(D)=0$ implies $P_{X_\RomanNumeralCaps{1}}^{s_2}(D)=0$ for any measurable $D$ and measures $P_{X_\RomanNumeralCaps{1}}^{r_2}, P_{X_\RomanNumeralCaps{1}}^{s_2}$ corresponding to $B$.
Therefore, $P_{X_\RomanNumeralCaps{1}^{'}}^{r_1}(D)=0$ implies $P_{X}^{s_2}(D)=0$.
\end{proof}

 \textbf{Classical (Expert-based)} data augmentation is often defined by domain experts who introduce simple transformations such as noise adding, cropping and swapping formalized above and denoted $A_{N}, A_{C}, A_{S}$, respectively.
    Following Lemma~\ref{lem:composition} the combination can be formalized by
    \begin{align}
        \label{ref:classical_augmentation}
        A:=A_{N}\circ A_{C}\circ A_{S}. 
    \end{align}

    \textbf{GAN}\footnote{A label-conditional GAN is intended here.} augmentation can be formalized as composition of class-specific data augmentations by interpreting the GAN training on the sample $S$ as part of the augmentation. In the setting of~\cite{goodfellow2014generative}, the output measure of the GAN equals the original measure of $S$ under general assumptions and therefore dominates the measure of $S$.
    
    \textbf{Mixup}~\cite{zhang2017mixup} data augmentation transforms the data by

    \begin{equation}
        \label{eq:mixup}
        \begin{split}
            A(S)= \big( &\big((1-M_1)\cdot \x_{i'} + M_1\cdot \x_{j'},(1-M_1)\cdot l(\x_{i'}) + M_1\cdot l(\x_{j'})\big),\ldots, \\ &  \big((1-M_r)\cdot \x_i + M_r\cdot \x_j,(1-M_r)\cdot l(\x_i) + M_r\cdot l(\x_j)\big)\big),
        \end{split}
    \end{equation}
    where $ i,i',j,j'\in\{1,\ldots,s\},\ i\neq i'\neq j\neq j'$ and $M_1,\ldots,M_r\sim \text{Beta}(\alpha, \alpha)$ for $\alpha>0$.

\subsection{Training Setup}
\label{subsec:training_setup}
\subsubsection{Image classification}
\label{subsec:image_classification_setup}

Image classification experiments are carried out on CIFAR10~\cite{krizhevsky2009learning} using a ResNet50~\cite{he2016deep}.
The ResNet model was trained using SGD and weight-decay with penalty coefficient of $5e-4$. 
Each classifier was trained for 200 epochs and learning rate schedule was used with initial value of $0.1$, which was reduced twice by a factor of 10 every 80 epochs.
This resulted in the best classification error of $5.39\pm0.10\%$ on the test set using classical data augmentation with $P_{Aug}=0.8$.

\subsubsection{Data Augmentation}
\label{subsec:data_augmentation_setup}

During training, using a stochastic data loader, samples are drown from the augmented set with probability $P_{Aug}$, and $1-P_{Aug}$ from the original training set.

For classical augmentation, the random rotation is done by up to 2 degrees; 
brightness, contrast and saturation distortion values are uniformly chosen from $\left[0.75, 1.25\right]$; 
and random cropping is applied after padding 4 pixels to each side of the image.

For GAN augmentations, an `augmented dataset' is created for each of the GAN models, using GAN-generated images.
The size of this dataset is 200 times larger than the training set of CIFAR10, as the models are trained for 200 epochs.
For each of the GAN-augmentations, given every training sample, 200 samples are generated and used as an `augmented' version of the chosen training sample. The generator is conditioned on the label of each training sample.

For Mixup and Classical augmentations, an `augmented dataset' is created with the same size as the training-set of CIFAR10 (50k): for each training sample, 1 augmented data is created using the augmentation method.
For models trained with Mixup and Classical augmentations, the augmented samples are then chosen from the created dataset with probability $\frac{1}{200}$, or otherwise are created on-the-fly\footnote{This is done to reduce loading time and speedup experiments, as Classical and Mixup augmentations are faster if created on the fly.}.

The `augmented datasets' were used in the evaluations of models with prediction-change stress around the augmented data, and the influence scores of the augmented data.

\subsubsection{Additional Discussions about Data Augmentation}
\label{subsec:data_augmentation_setup_additional_info}

In our experiments, each augmentation method is applied alone.
Although the results in~\cite{zhang2017mixup} suggests that Mixup improves adversarial robustness (against FGSM~\cite{goodfellow2014explaining} attacks), the models in~\cite{zhang2017mixup} were trained with a combination of Mixup and Classical augmentation.

\subsection{Attack Setup}
\label{appx:attack_setup}
We carry out 4 different untargeted PGD attacks with $l2$ norm.
Additionally, we apply 6  untargeted PGD attacks with $l_{inf}$ norm.
The parameters of these attacks are provided in Table~\ref{tab:pgd_params}.

\begin{table}[ht]
\caption{PGD attack parameters used in the experiments.}
\label{tab:pgd_params}
\centering
\begin{tabular}{lcccc}
$PGD_{\epsilon}$&$PGD_{\alpha}$&\multicolumn{2}{c}{Iterations} &norm\\ \hline
0.25 &  0.5&10 & 100 &$l_2$\\ 
0.5 & 0.1 &10 & 100 &$l_2$\\ \hline
0.03 & 0.006 &10 & 100 &$l_{inf}$\\ 
0.05 & 0.01& 10 & 100 &$l_{inf}$\\ 
0.1 &0.02 &  10& 100 &$l_{inf}$\\ \hline
\end{tabular}
\end{table}

\subsection{GAN Setup}
\label{appx:gan_setup}
\begin{figure}[h]
\centering
    \begin{subfigure}[a]{.49\textwidth}
           \centering
           \includegraphics[width=\linewidth]{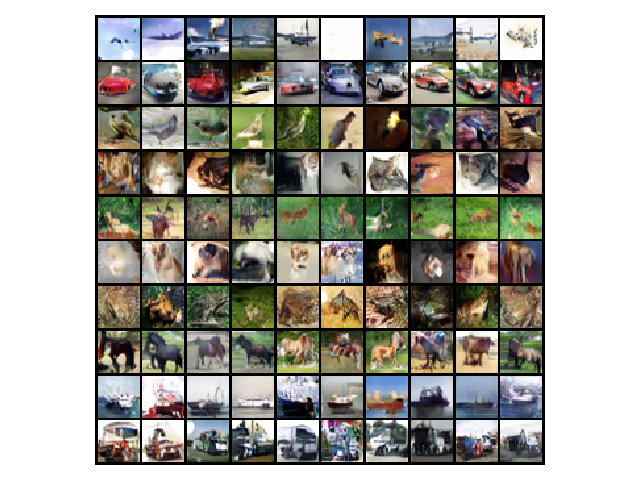}
           \caption{NS GAN.}
           \label{subfig:ns_samples}
    \end{subfigure}
    \hfill
    \begin{subfigure}[a]{.49\textwidth}
           \centering
           \includegraphics[width=\linewidth]{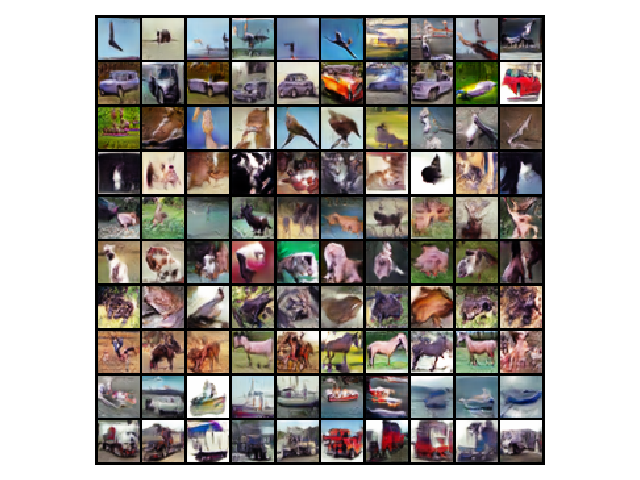}
           \caption{WGP GAN.}
           \label{subfig:wgp_samples}
    \end{subfigure}
\caption{Random generated samples from the NS and WGP GANs, conditioned on different labels. The conditioning labels in each row are fixed. }
\label{fig:gan_samples}
\end{figure}

\begin{table}[t]
\caption{FID for WGP and NS GANs on CIFAR10.}
\label{tab:fid}
\centering
\begin{tabular}{lcc}
    & WGP   & NS \\ \hline
FID & 20.11 & 18.30 \\ \hline
\end{tabular}
\end{table}

We use a GAN that incorporates convolutional layers with residual connections in both generator and discriminator.
The TTUR learning-rates~\cite{heusel2017gans} were used for the generator and the discriminator.
A 100-dimensional noise was concatenated with a 10-dimensional one-hot encoded labels as the input to the generator.
The discriminator consist of a model with 2 outputs: one for discriminating between the real and fake, and another for classifying the image into 10 classes.
The classification and GAN objective were jointly optimized.
The GANs were trained for 120 epochs.
No data augmentation was used in training of the GAN models.
The GAN architectures are detailed in Section~\ref{appx:network_architectures}.
The trained GANs are evaluated with the `Fr\'{e}chet Inception Distance' (FID), and the results of this evaluation can be found in Table~\ref{tab:fid}.

\subsection{Extended Results for Adversarial Attacks}
\label{appx:extended_empirical_results}
\begin{figure}[ht]
\centering
\includegraphics[width=\linewidth]{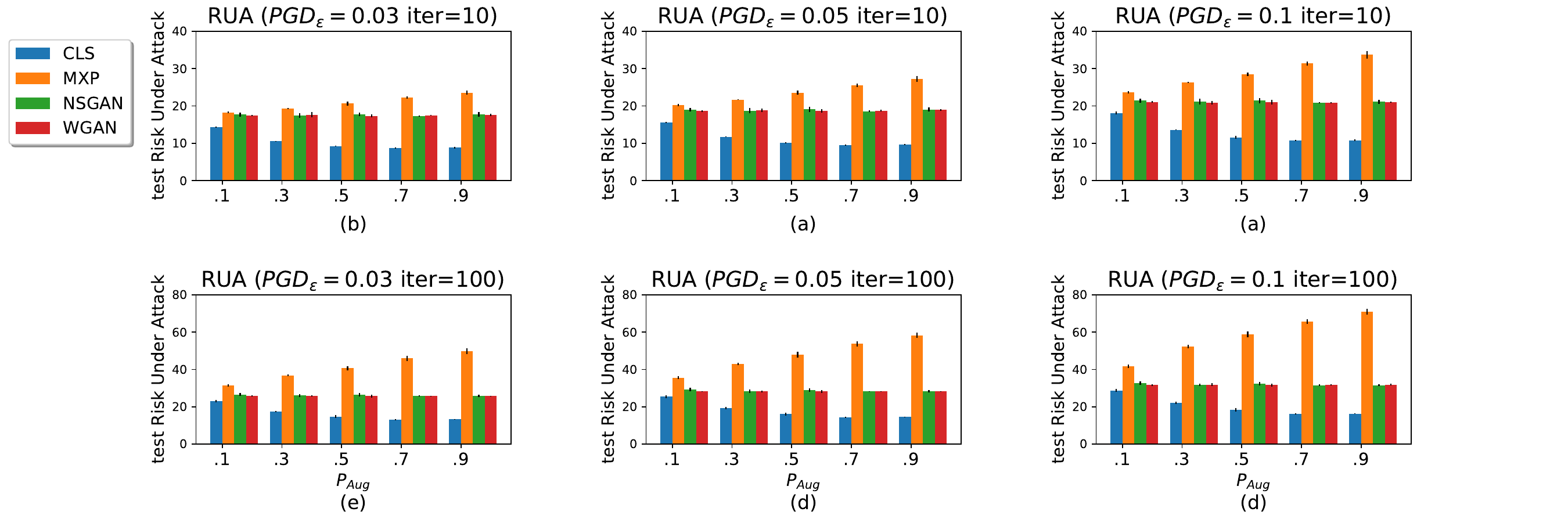}
\caption{
Risk Under Adversarial Attack (RUA) for PGD attacks with $l_{inf}$ norm for different augmentation methods on the test set of CIFAR10.
}
\label{fig:rua_linf}
\end{figure}

In Figure~\ref{fig:rua_linf} we present the results of Risk Under Attack with PGD, using $l_{inf}$ norm, for different iterations and $PGD_\epsilon$ values.
As can be seen, similar to the results in Figure~\ref{fig:clf_results} Mixup achieves the highest risk under adversarial attack in all cases.
It can additionally be observed that while risk under adversarial attack increases as the augmentation probability rises for models trained with Mixup, the models trained with classical augmentation result in lower risk as the augmentation probability increases.
These observations are also on par with the results presented in Section~\ref{subsec:clferr_rua}.

\subsection{Extended Results for Influence Analysis}
\label{appx:extended_influence_results}
In this section we provide the extended results for the influence scores, the loss distribution, and the average gradient-norm distribution of the augmented, and non-augmented samples.

\subsubsection{The Distribution of the Average Gradient-Norm and the Loss}
\begin{figure}[ht]
	\begin{subfigure}[1a]{.5\linewidth}
	\centering
	\includegraphics[width=\textwidth]{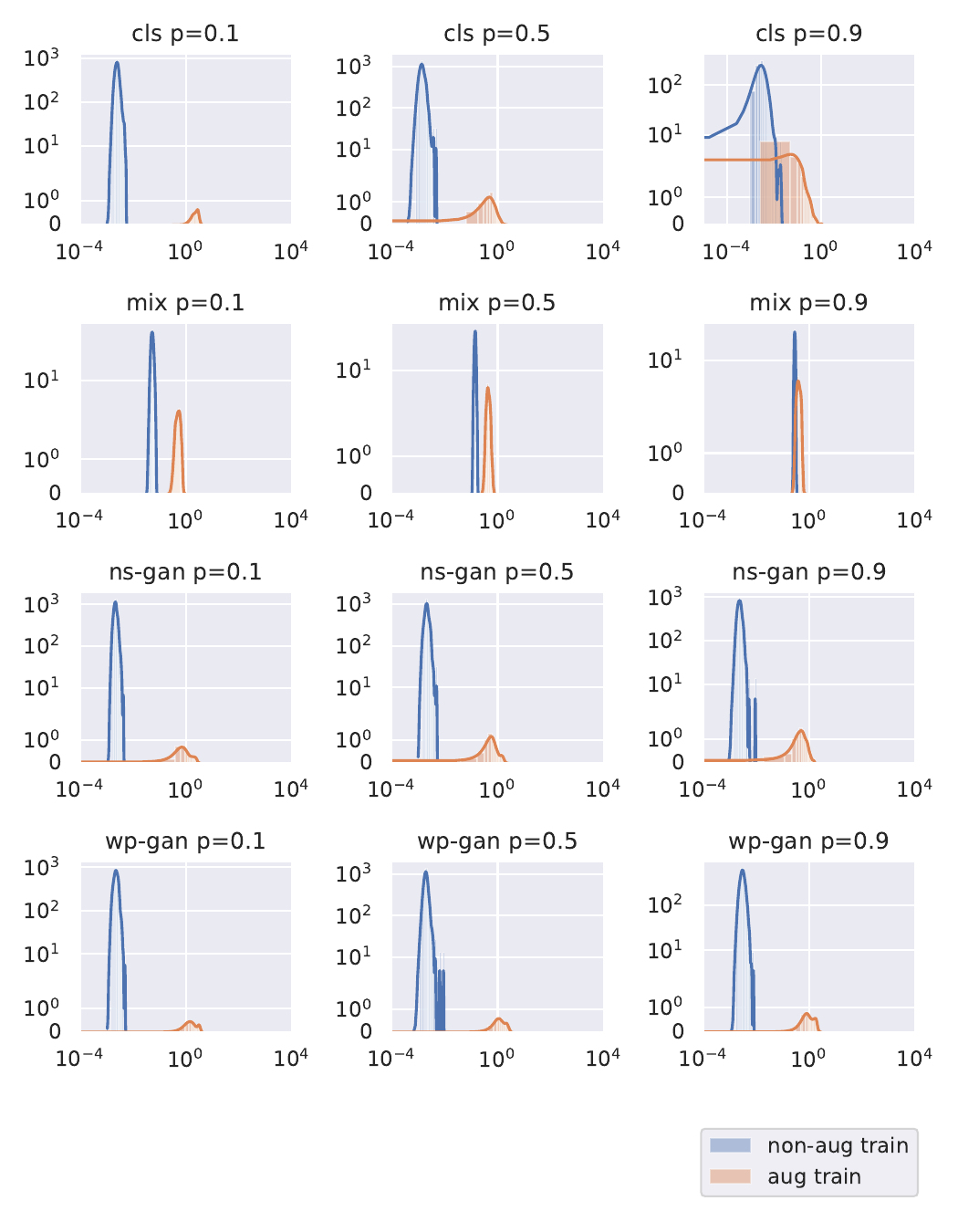}
	\end{subfigure}
	\hfil
	\begin{subfigure}[1b]{.5\linewidth}
	\includegraphics[width=\textwidth]{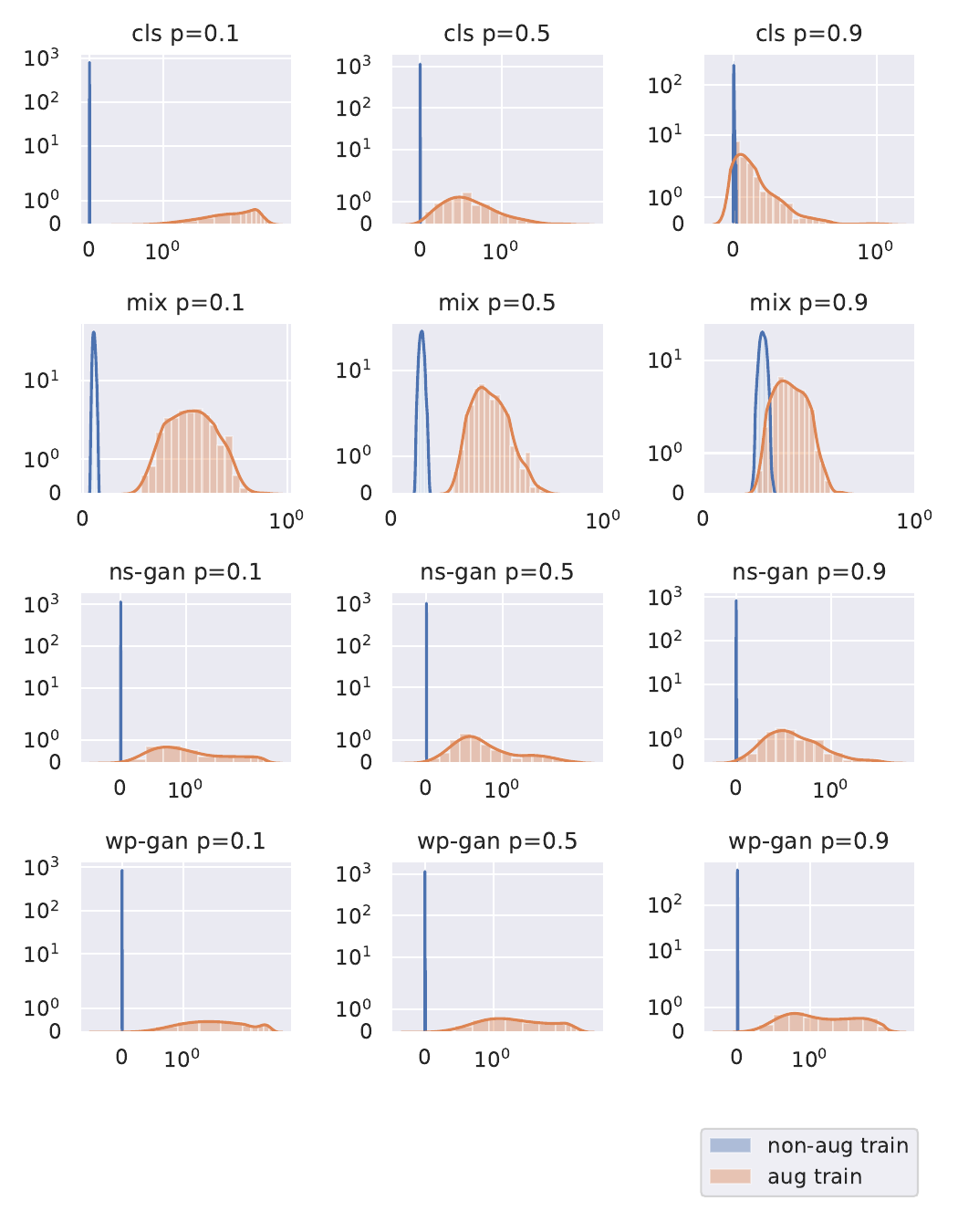}
	\end{subfigure}
\caption{Distribution of the average gradient-norm for augmented and non-augmented data.
The plot on the left has a logarithmically scaled X-axis, and a symmetrically logarithmically scaled Y-axis.
In the plot on the right, both X and Y axes are scaled symmetrically logarithmic for a better view.}
\label{fig:grad_distribution}
\end{figure}

\begin{figure}[ht]
	\begin{subfigure}[1a]{.5\linewidth}
	\centering
	\includegraphics[width=\textwidth]{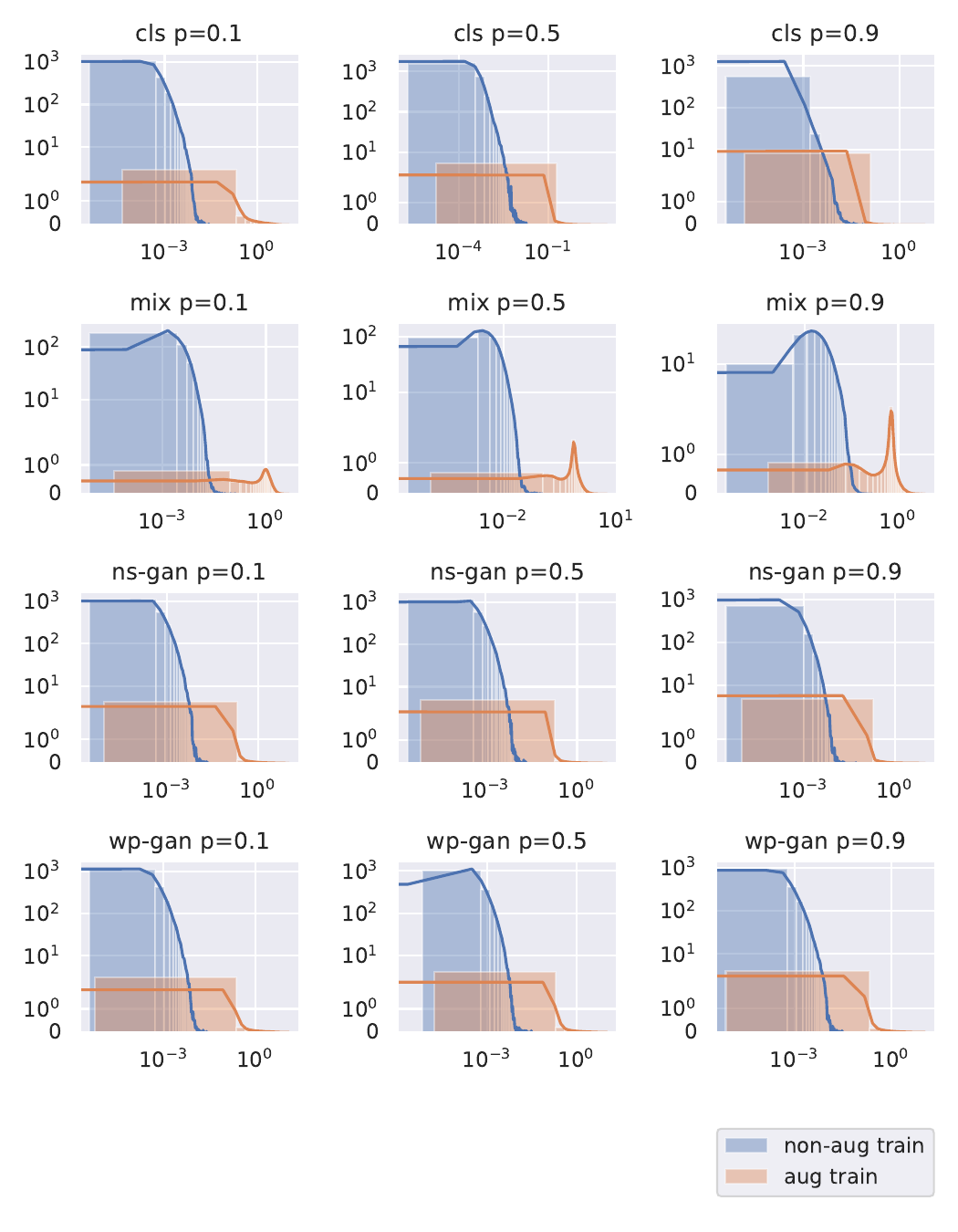}
	\end{subfigure}
	\hfil
	\begin{subfigure}[1b]{.5\linewidth}
	\includegraphics[width=\textwidth]{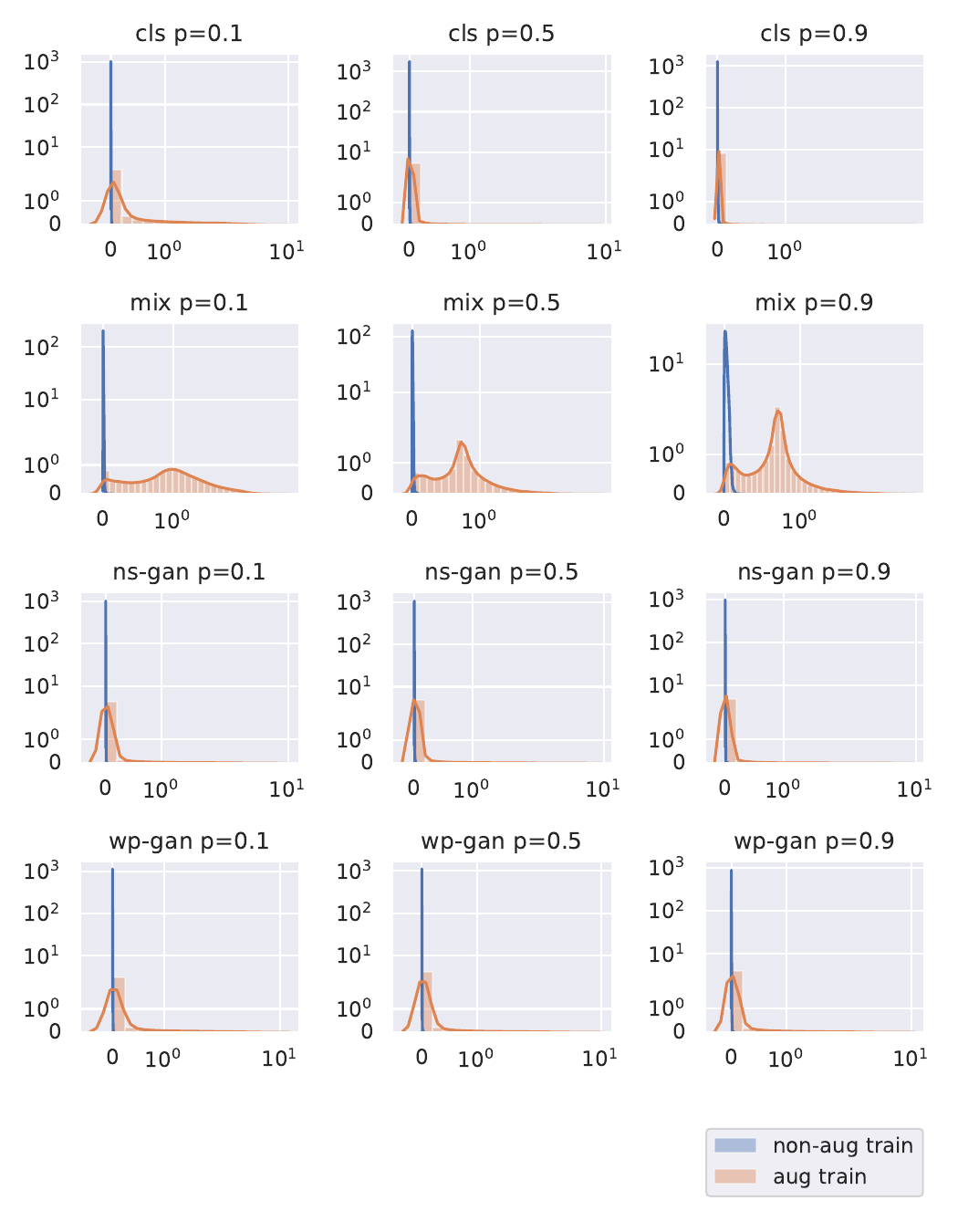}
	\end{subfigure}
\caption{Distribution of loss for augmented and non-augmented data.
The plot on the left has a logarithmically scaled X-axis, and a symmetrically logarithmically scaled Y-axis.
In the plot on the right, both X and Y axes are scaled symmetrically logarithmic for a better view.}
\label{fig:loss_distribution}
\end{figure}

The distribution of the `average gradient-norm' for augmented, and the non-augmented data, on various models are provided in Figure~\ref{fig:grad_distribution}.
The average gradient-norm for a minibatch of training examples is calculated by the $l_2$ norm of the gradient of the loss w.r.t the model parameters, averaged over all parameters.
Additionally, the distribution of the loss for augmented, and non-augmented data, can be found in Figure~\ref{fig:loss_distribution}.
All models used for the loss and the average gradient-norm plots are fully-trained.

As can be seen, the distribution values for both loss and average gradient-norms suggest that the value ranges in the augmented data, are higher than the non-augmented data, across all models.
The relatively lower average gradient-norm in the non-augmented data suggests that the parameters of the model do not change in the same order when trained on the non-augmented data, in comparison with the augmented data.
In contrast, when trained on augmented samples, models have a higher average gradient-norm, which results in a relatively higher change in the parameters of the models.

This observation suggests that models are more affected by the augmented data, than the non-augmented data.
These results are also aligned with the influence score results presented in Section~\ref{subsec:is}, which stated the models are influenced by the augmented data more than the non-augmented data.

\subsubsection{The Distribution of the Influence Scores}

\begin{figure}[ht]
    
    \begin{subfigure}[1a]{.24\linewidth}
          \includegraphics[width=1.\linewidth]{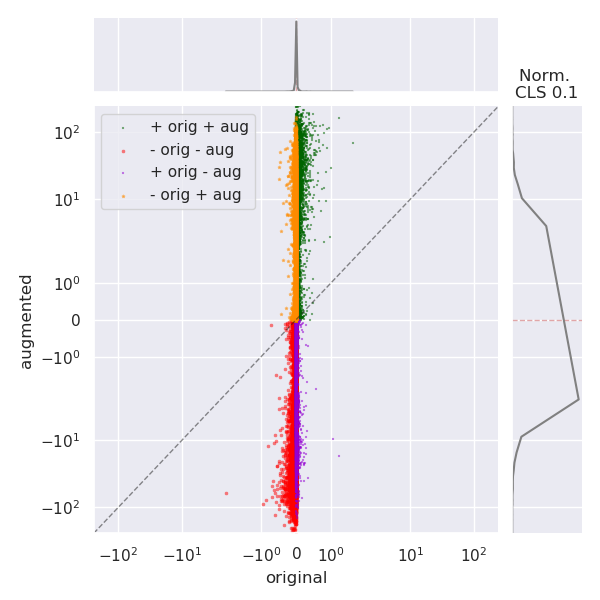}

    \end{subfigure}
    \hfil
    \begin{subfigure}[2a]{.24\linewidth}
        
          \includegraphics[width=1.\linewidth]{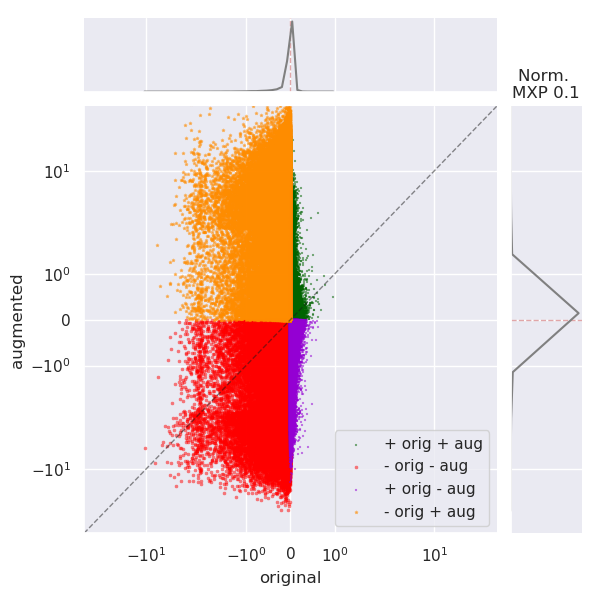}

    \end{subfigure}
    \hfil
    \begin{subfigure}[3a]{.24\linewidth}
        
          \includegraphics[width=1.\linewidth]{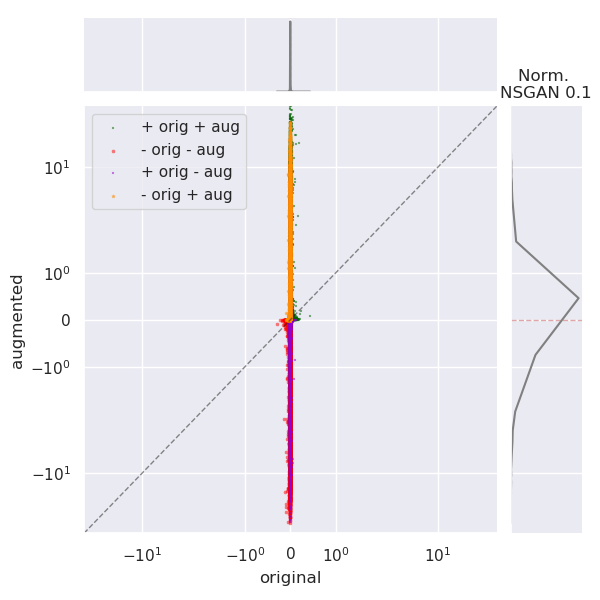}

    \end{subfigure}   
    \hfil
    \begin{subfigure}[4a]{.24\linewidth}
        
          \includegraphics[width=1.\linewidth]{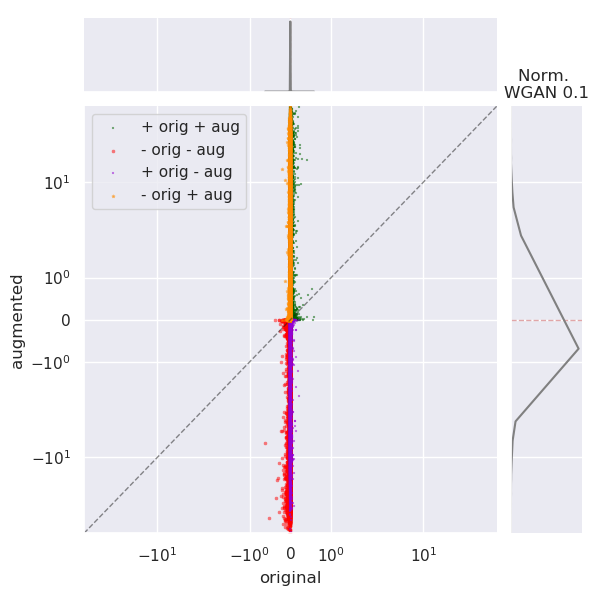}

    \end{subfigure}
    
    \vspace{1mm}
    \begin{subfigure}[1b]{.24\linewidth}
          \includegraphics[width=1.\linewidth]{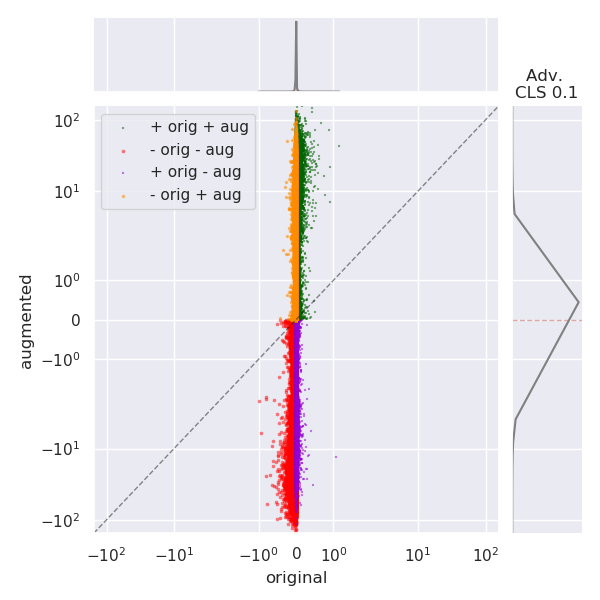}
    \end{subfigure}
    \hfil
    \begin{subfigure}[2b]{.24\linewidth}
        
          \includegraphics[width=1.\linewidth]{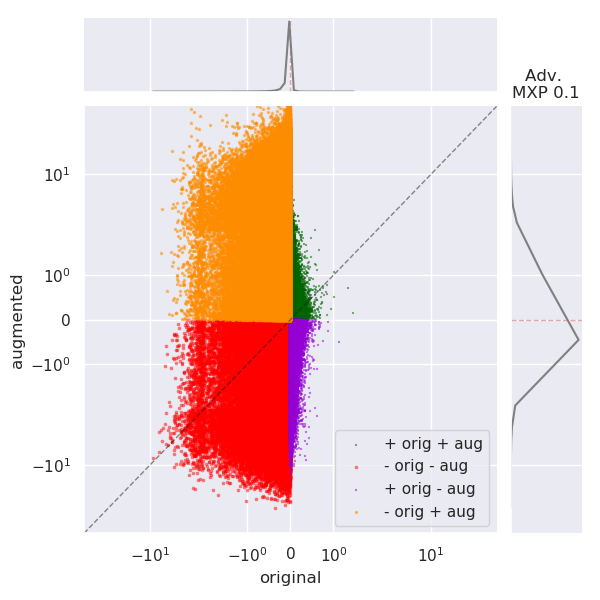}

    \end{subfigure}
    \hfil
    \begin{subfigure}[3b]{.24\linewidth}
        
          \includegraphics[width=1.\linewidth]{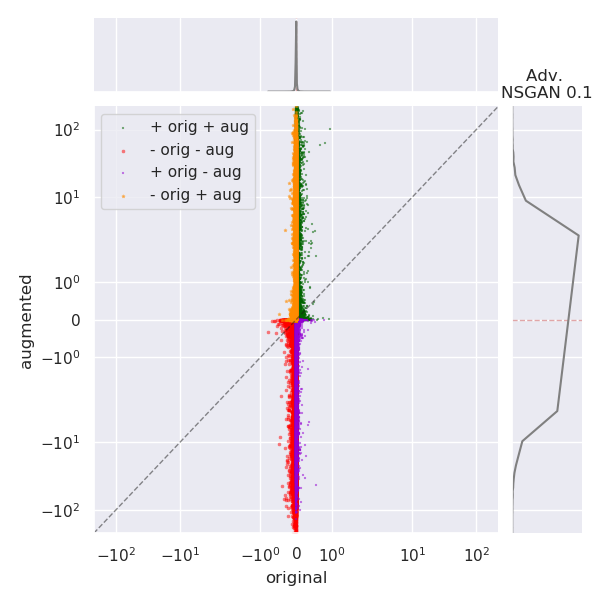}

    \end{subfigure}   
    \hfil
    \begin{subfigure}[4b]{.24\linewidth}
        
          \includegraphics[width=1.\linewidth]{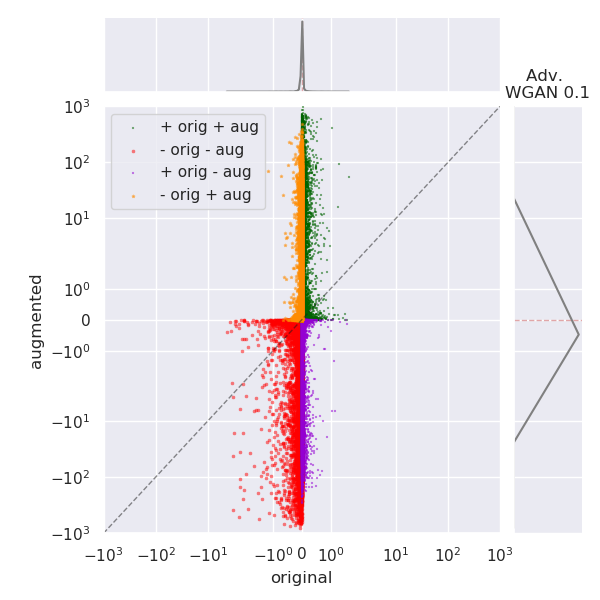}

    \end{subfigure}
    \vspace{1mm}

    
    \begin{subfigure}[1c]{.24\linewidth}
        
          \includegraphics[width=1.\linewidth]{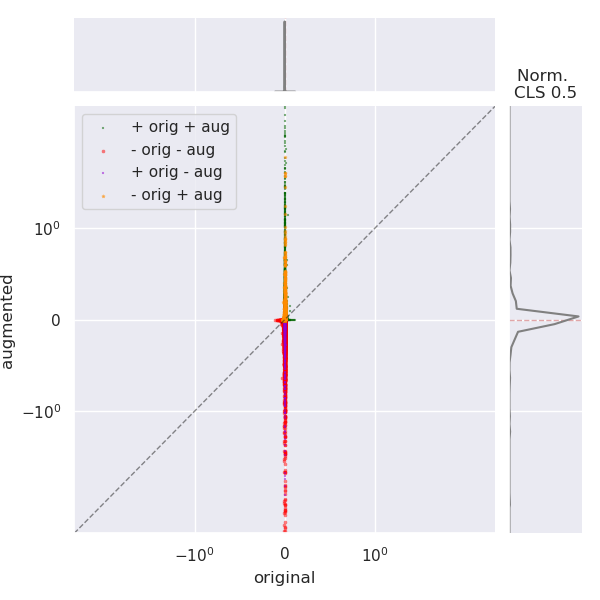}

    \end{subfigure}
    \hfil
    \begin{subfigure}[2c]{.24\linewidth}
        
          \includegraphics[width=1.\linewidth]{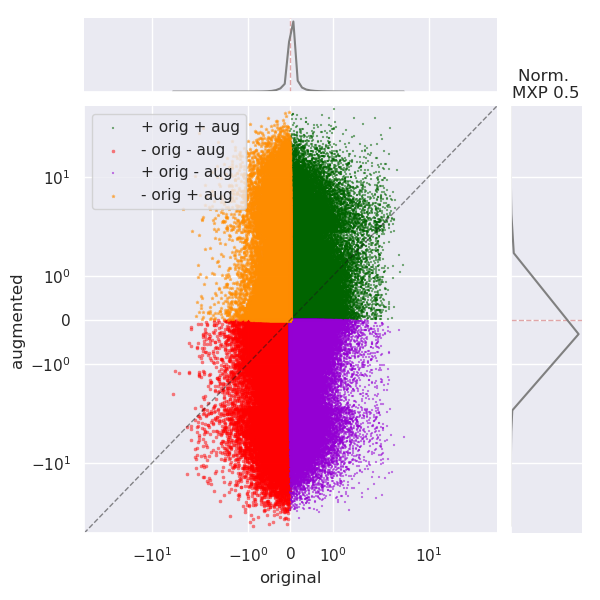}

    \end{subfigure}
    \hfil
    \begin{subfigure}[3c]{.24\linewidth}
        
          \includegraphics[width=1.\linewidth]{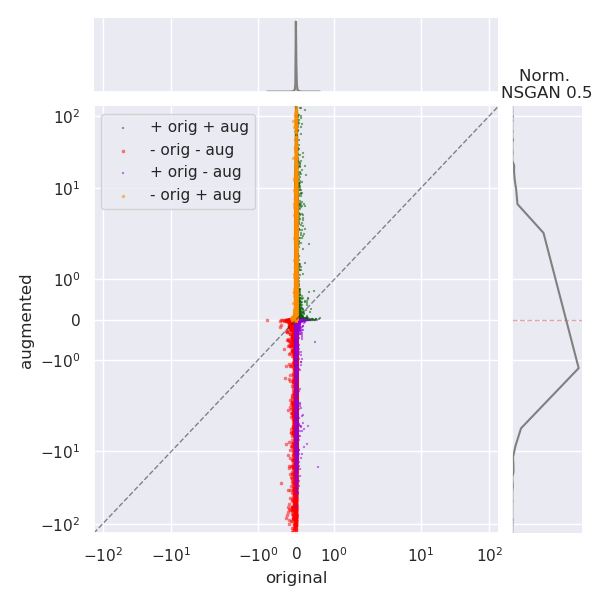}

    \end{subfigure}   
    \hfil
    \begin{subfigure}[4c]{.24\linewidth}
        
          \includegraphics[width=1.\linewidth]{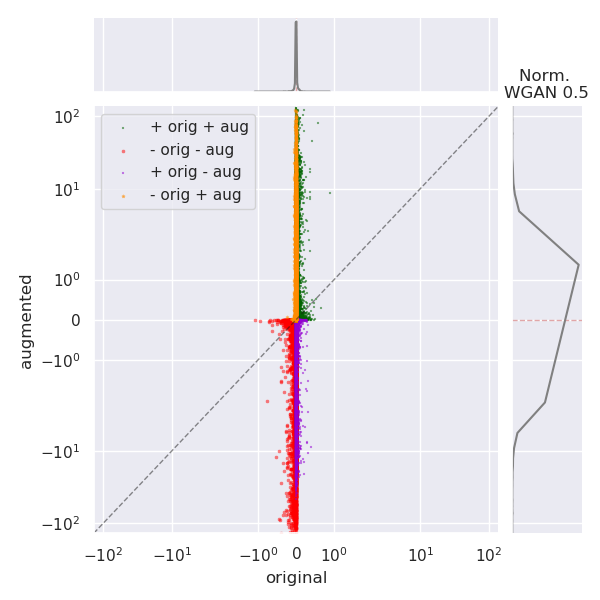}

    \end{subfigure}
    
    \vspace{1mm}
    \begin{subfigure}[1d]{.24\linewidth}
        
          \includegraphics[width=1.\linewidth]{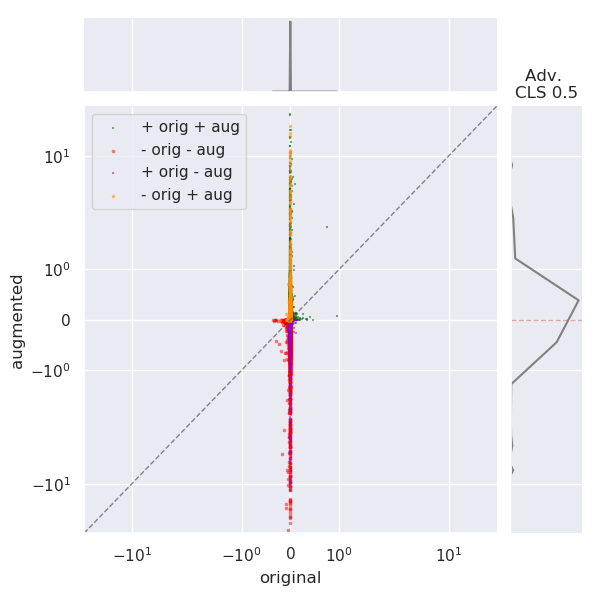}

    \end{subfigure}
    \hfil
    \begin{subfigure}[2d]{.24\linewidth}
        
          \includegraphics[width=1.\linewidth]{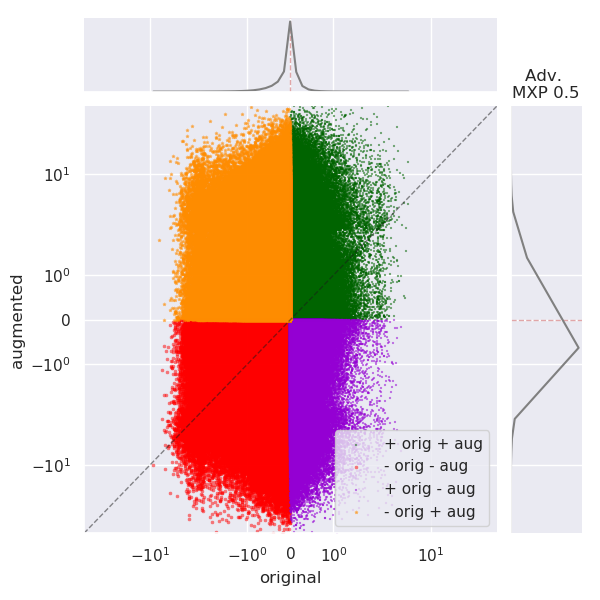}

    \end{subfigure}
    \hfil
    \begin{subfigure}[3d]{.24\linewidth}
        
          \includegraphics[width=1.\linewidth]{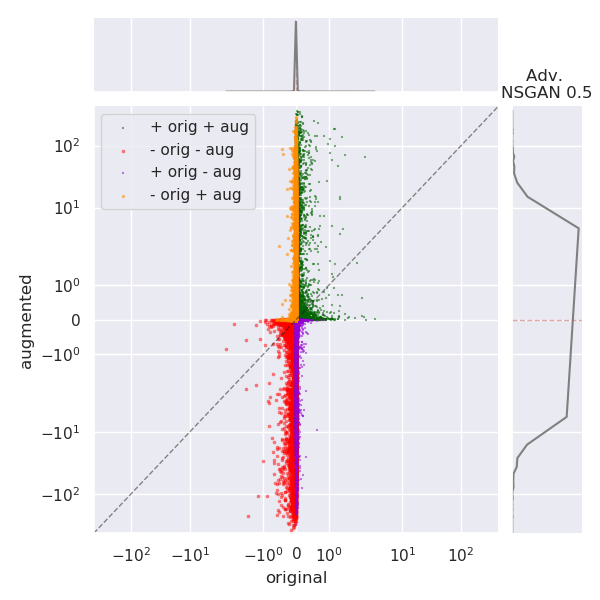}

    \end{subfigure}   
    \hfil
    \begin{subfigure}[4d]{.24\linewidth}
        
          \includegraphics[width=1.\linewidth]{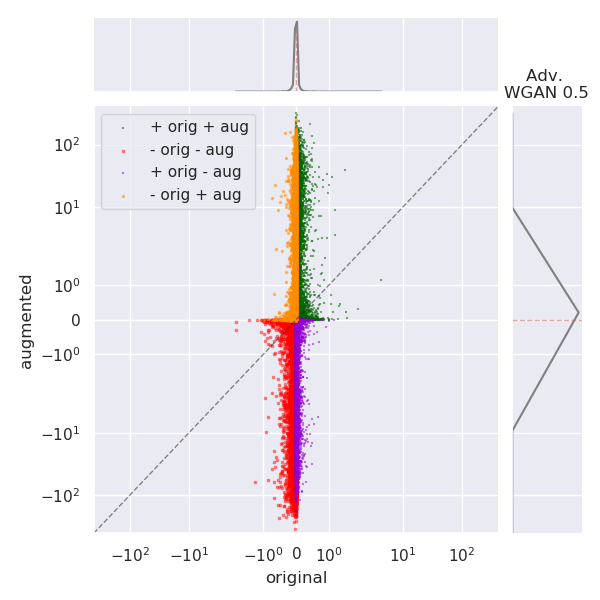}

    \end{subfigure}
    \vspace{1mm}

    
    \begin{subfigure}[1e]{.24\linewidth}
        
          \includegraphics[width=1.\linewidth]{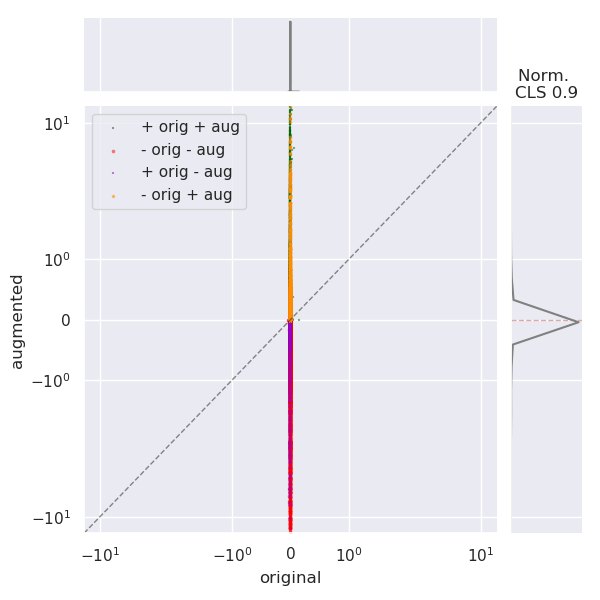}

    \end{subfigure}
    \hfil
    \begin{subfigure}[2e]{.24\linewidth}
        
          \includegraphics[width=1.\linewidth]{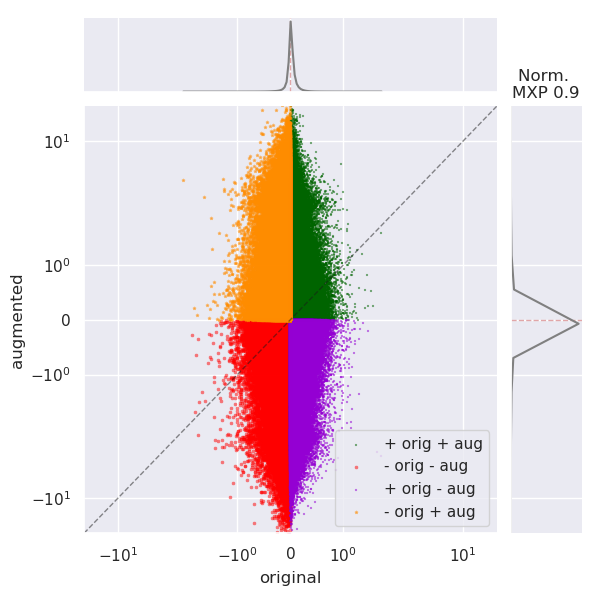}

    \end{subfigure}
    \hfil
    \begin{subfigure}[3e]{.24\linewidth}
        
          \includegraphics[width=1.\linewidth]{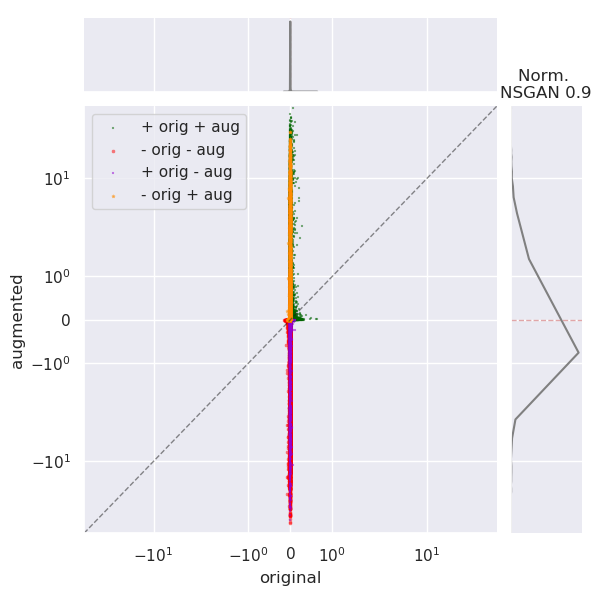}

    \end{subfigure}   
    \hfil
    \begin{subfigure}[4e]{.24\linewidth}
        
          \includegraphics[width=1.\linewidth]{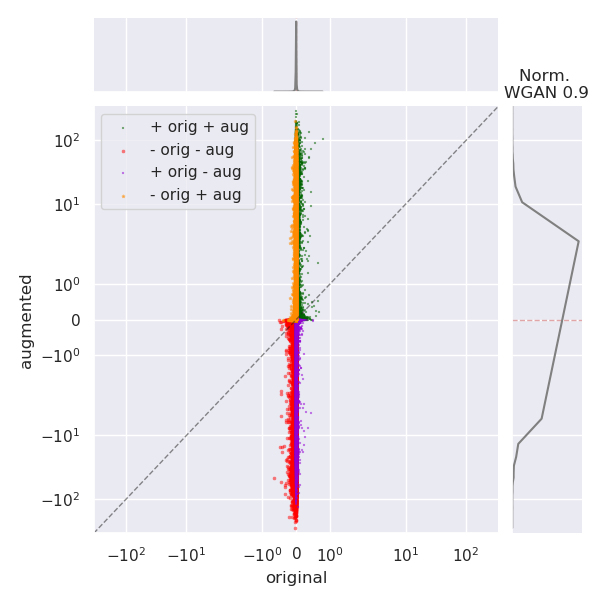}

    \end{subfigure}
    
    \vspace{1mm}
    \begin{subfigure}[1f]{.24\linewidth}
        
          \includegraphics[width=1.\linewidth]{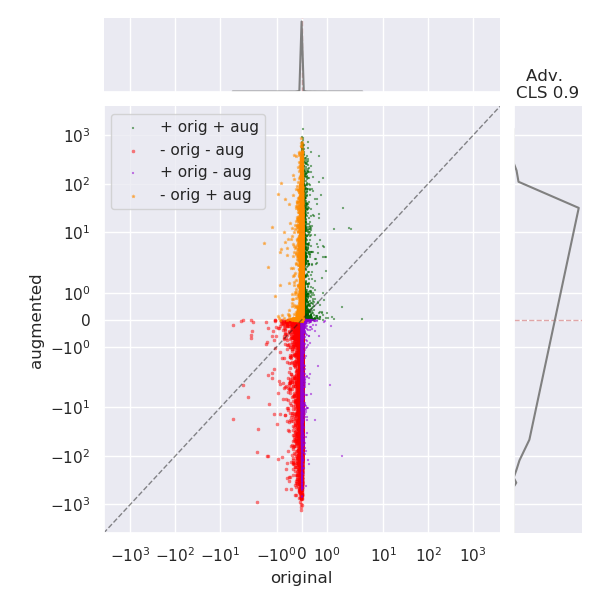}

    \end{subfigure}
    \hfil
    \begin{subfigure}[2f]{.24\linewidth}
        
          \includegraphics[width=1.\linewidth]{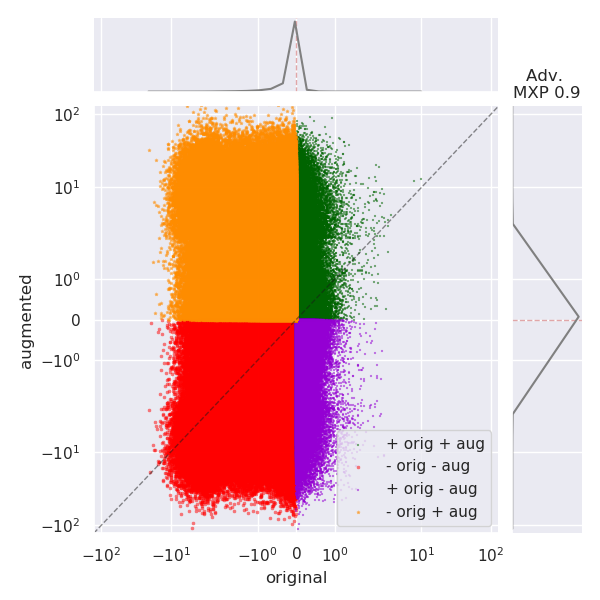}

    \end{subfigure}
    \hfil
    \begin{subfigure}[3f]{.24\linewidth}
        
          \includegraphics[width=1.\linewidth]{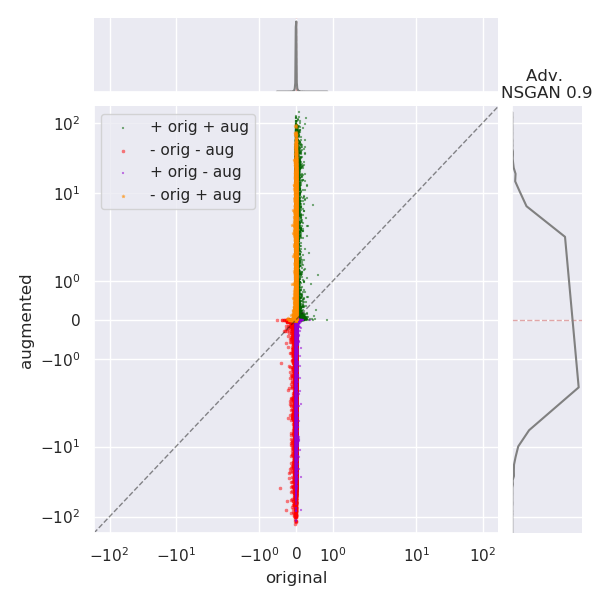}

    \end{subfigure}   
    \hfil
    \begin{subfigure}[4f]{.24\linewidth}
        
          \includegraphics[width=1.\linewidth]{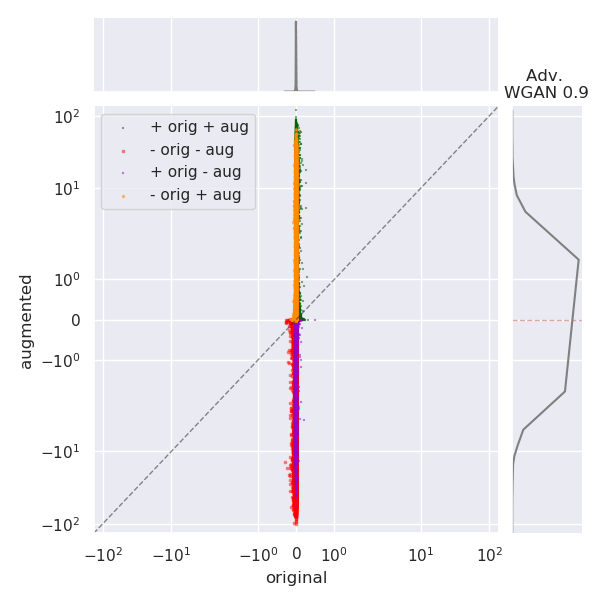}

    \end{subfigure}
    \vspace{1mm}
\caption{Comparison between the positive and negative influence values of all original data vs. their augmented counterparts for different augmentation methods, when predicting on the real and the adversarial test set of CIFAR10.
The side histograms represent the marginal distribution of influence values for augmented and original (non-augmented).
Both axis have a symmetrical logarithmic scale, with equal length.
The black dashed line represent the line $X=Y$.
The red dashed lines in the side plots, represent the zero value reference.
}
\label{fig:influence_scatter_results}
\end{figure}

In Figure~\ref{fig:influence_scatter_results}, we provide the scatter plots of the influence scores of augmented and non-augmented examples, when predicting on the test set, and their adversarial counterparts.
The marginal histograms for all augmented and non-augmented influence scores are also provided in the side plots.
For a better view, we separate the colours for the positive ($+$) and negative ($-$) influence values for augmented (aug) and non-augmented (orig) training data.
Each point on the scatter plot, represents the influence score of a training example, for a model when predicting on an unseen (test or adversarial) example.
The X value of the point, represents the influence score of a non-augmented example, while its Y value shows the influence score of its augmented counterpart.

Looking at the diagonal black dashed line, we can see that the datapoints are \emph{not} scattered along this line, which means the influence of non-augmented samples are not similar to the influence of their augmented counterparts.
As can be seen, the influence from the augmented data is larger (in both positive and negative), compared to non-augmented data, as it was  previously reported in Figure~\ref{fig:influence_dist_results}.

We can additionally observe that in the case of Mixup, the influence scores are more spread compared to other augmentations.
Looking at the vertical side histogram plots (right side), we can observe that in the majority of the models trained with Mixup, 
the distribution of the influence scores for the augmented data is shifted more to the negative side (bottom side of the red dashed line) when models predict on adversarial examples, in comparison with predicting on normal test examples.
As the negative influence values represent training data points that contribute to \emph{increase} the loss on a given unseen example, they have a negative influence on the model's prediction.
This observation suggests that the Mixup-augmented training examples have more negative influence when predicting on adversarial examples.
These results are also aligned with the risk under attack (see Figures~\ref{fig:clf_results},~\ref{fig:rua_linf},~\ref{fig:avg_stress_results}, and \ref{fig:linf_avg_stress_results}) which shows models trained with Mixup are more vulnerable to adversarial attacks.

\subsection{Extended Results for the Prediction-Change Stress}
\label{appx:extended_prediction_change_stress_results}
\subsubsection{Results with $l_2$ norm}
\label{appx:subsec:pce_l2}
\begin{figure}[hbt!]
\centering
    \begin{subfigure}[b]{.3\columnwidth}
           \centering
\includegraphics[width=1.\textwidth]{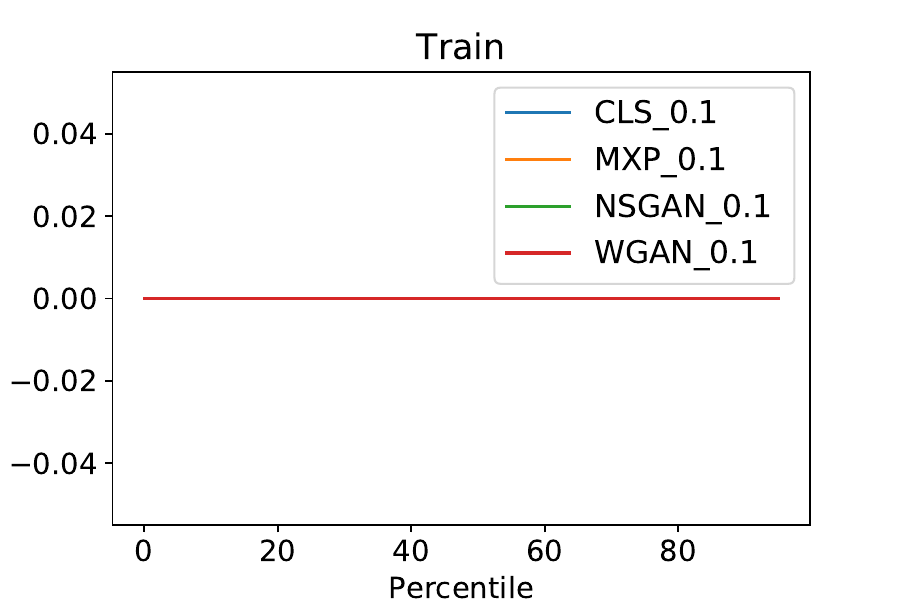}
            \caption{.}
            \label{subfig:}
    \end{subfigure}
    \hfil
    \begin{subfigure}[b]{.3\columnwidth}
           \centering
           \includegraphics[width=1.\textwidth]{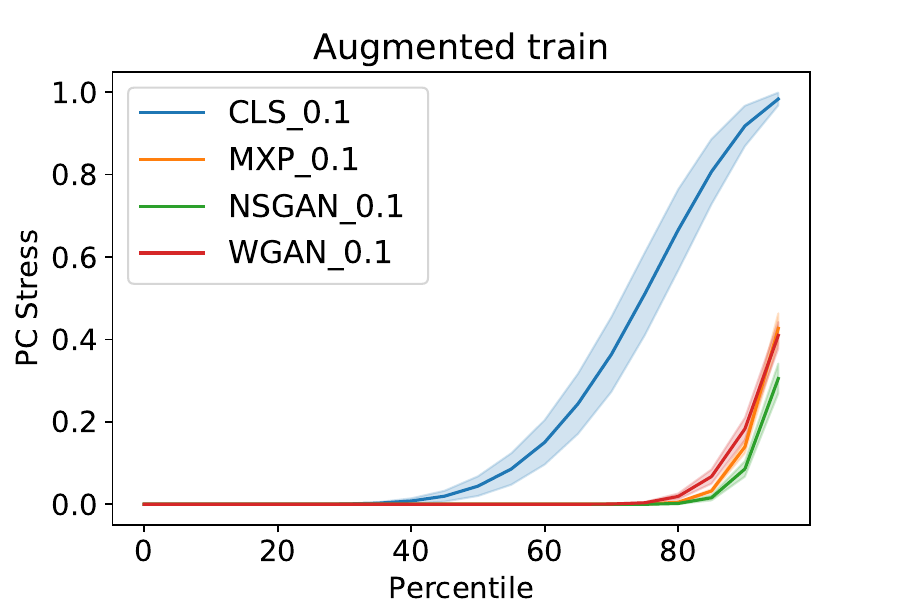}
            \caption{.}
            \label{subfig:}
    \end{subfigure}
    \hfil
    \begin{subfigure}[b]{.3\columnwidth}
           \centering
           \includegraphics[width=1.\textwidth]{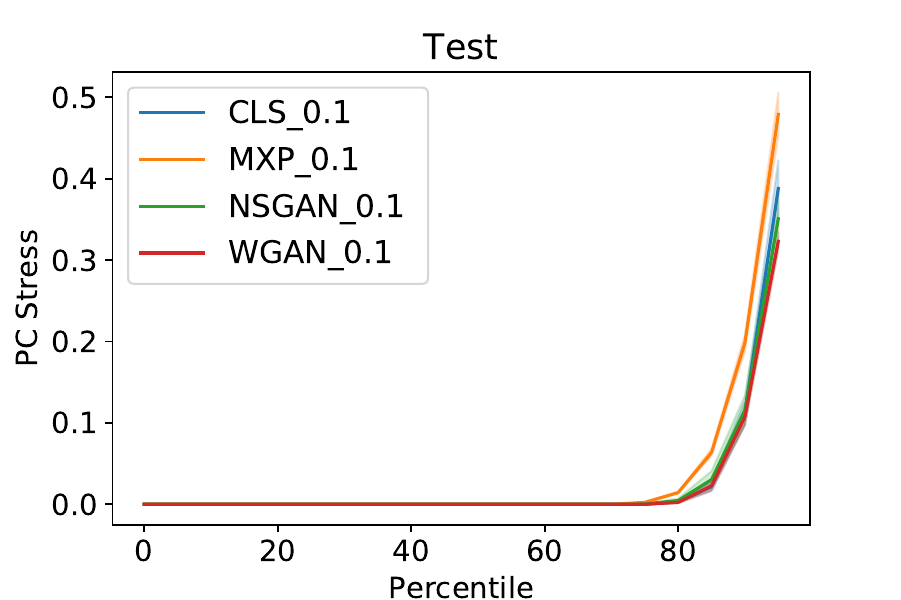}
            \caption{.}
            \label{subfig:}
    \end{subfigure}
    \vspace{1mm}
    \begin{subfigure}[b]{.3\columnwidth}
           \centering
           \includegraphics[width=1.\textwidth]{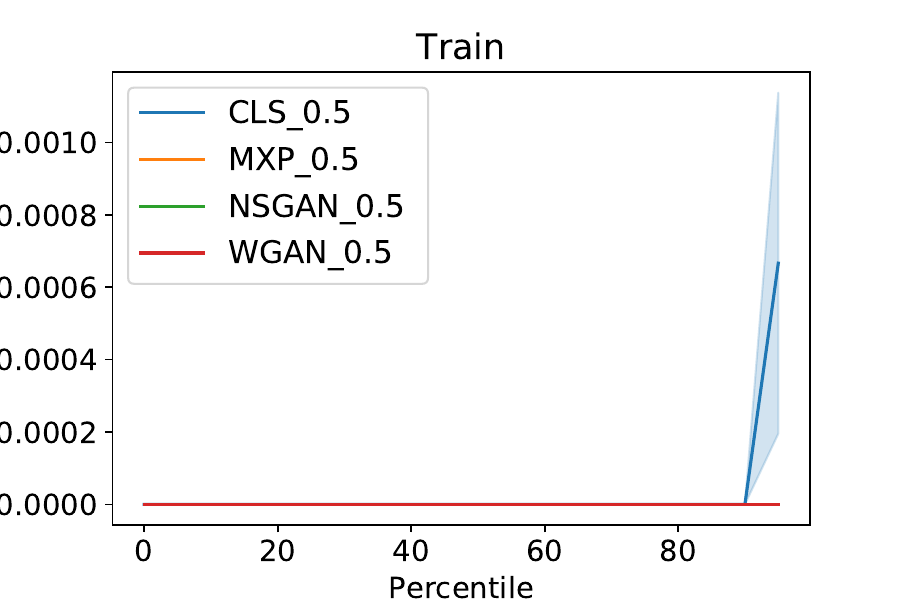}
            \caption{.}
            \label{subfig:}
    \end{subfigure}
    \hfil
    \begin{subfigure}[b]{.3\columnwidth}
           \centering
           \includegraphics[width=1.\textwidth]{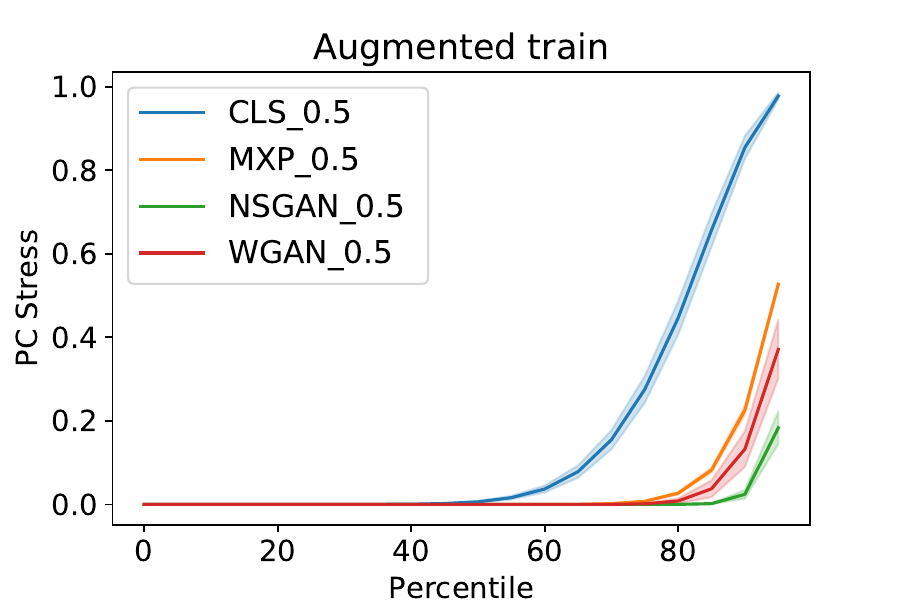}
            \caption{.}
            \label{subfig:}
    \end{subfigure}
  \hfil
    \begin{subfigure}[b]{.3\columnwidth}
           \centering
           \includegraphics[width=1.\textwidth]{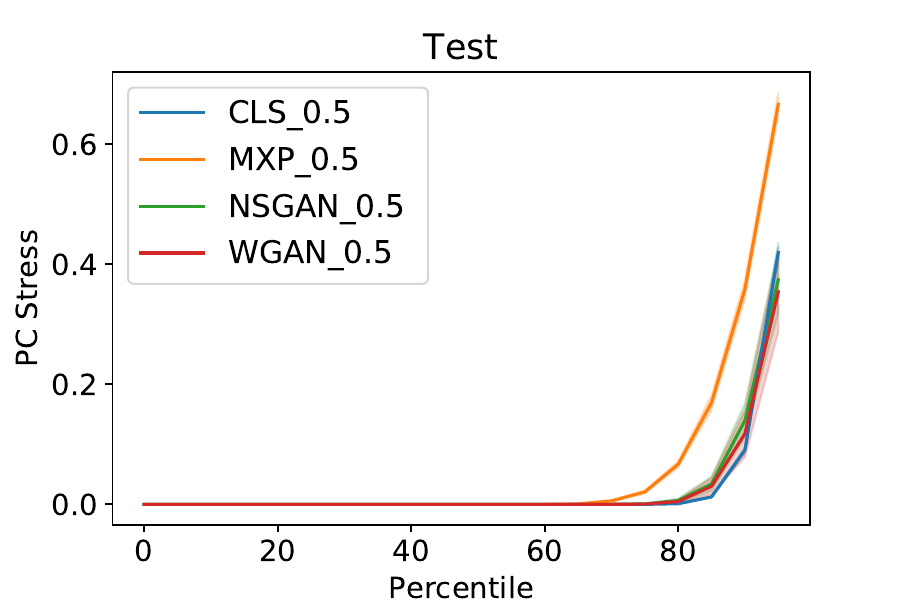}
            \caption{.}
            \label{subfig:}
    \end{subfigure}
    \vspace{1mm}
    \begin{subfigure}[b]{.3\columnwidth}
           \centering
           \includegraphics[width=1.\textwidth]{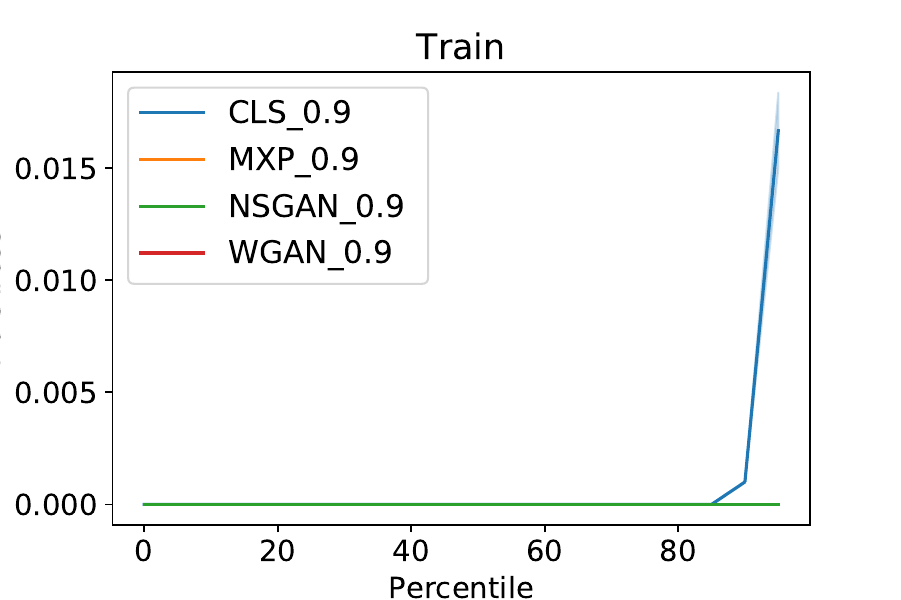}
            \caption{.}
            \label{subfig:}
    \end{subfigure}
    \hfil
    \begin{subfigure}[b]{.3\columnwidth}
           \centering
           \includegraphics[width=1.\textwidth]{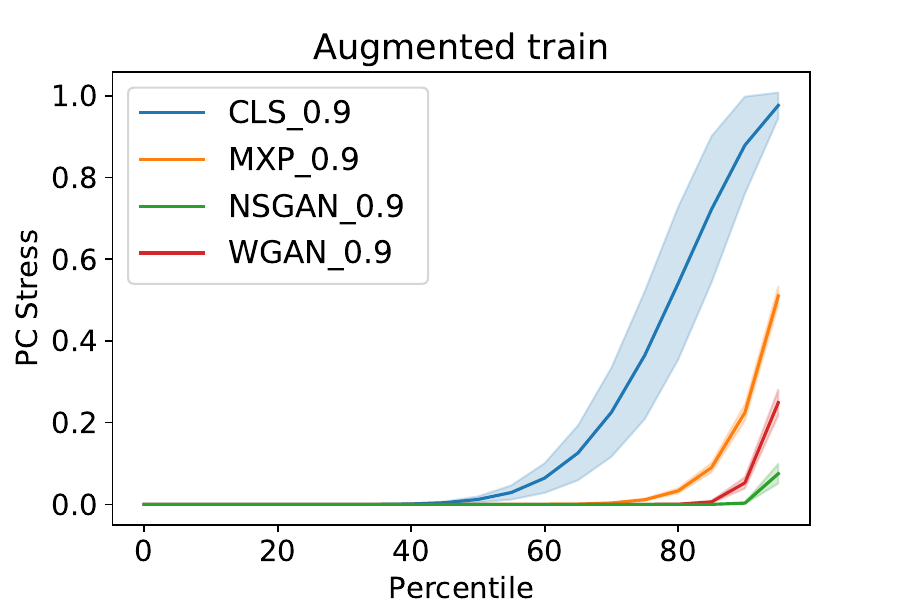}
            \caption{.}
            \label{subfig:}
    \end{subfigure}
    \hfil
    \begin{subfigure}[b]{.3\columnwidth}
           \centering
           \includegraphics[width=1.\textwidth]{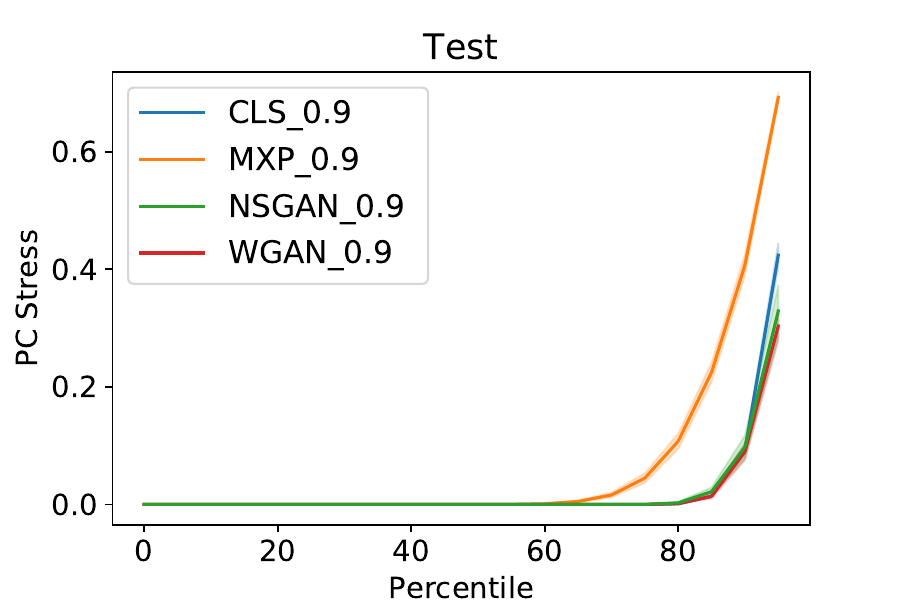}
            \caption{.}
            \label{subfig:}
    \end{subfigure}%
\caption{
Comparison between the distribution Prediction-Change Stress of models with different augmentations using $l_2$ norm, around original training, augmented training, and test set of CIFAR10, and for $\epsilon=1$.
First row: 0.1 augmentation.
Second row: 0.5 augmentation.
Third row: 0.9 augmentation.
First column: PCS around the original training set.
Second column: PCS around the augmented training set.
Third column: PCS around the test set.
}
\label{fig:stress_results}
\end{figure}

\begin{figure}[hbt!]
	\centering
	\includegraphics[width=1.03\textwidth]{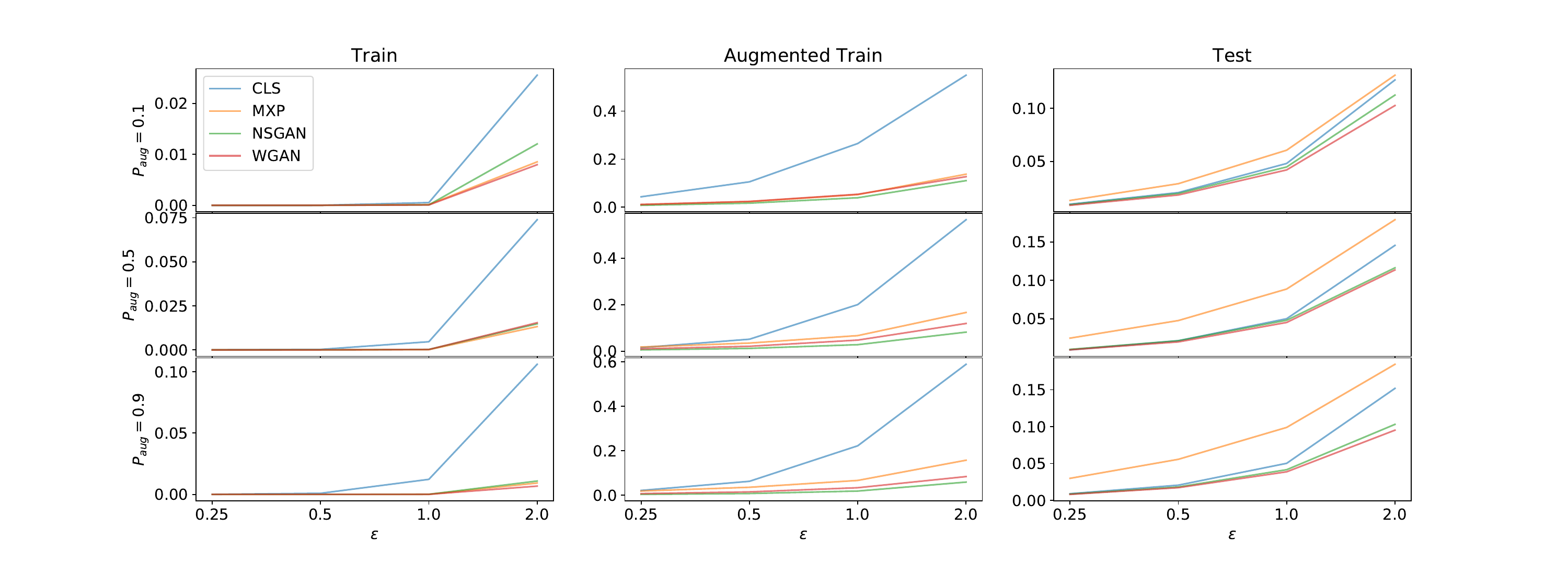}
	\caption{
		Comparison between the average Prediction-Change Stress of models with different augmentations using $l_2$ norm.
		The y-axis is the  average Stress calculated over all the original training, augmented training, and test set of CIFAR10.
	}
	\label{fig:avg_stress_per_epsilon}
\end{figure}

\begin{figure}[hbt!]
	\centering
	\includegraphics[width=1.03\textwidth]{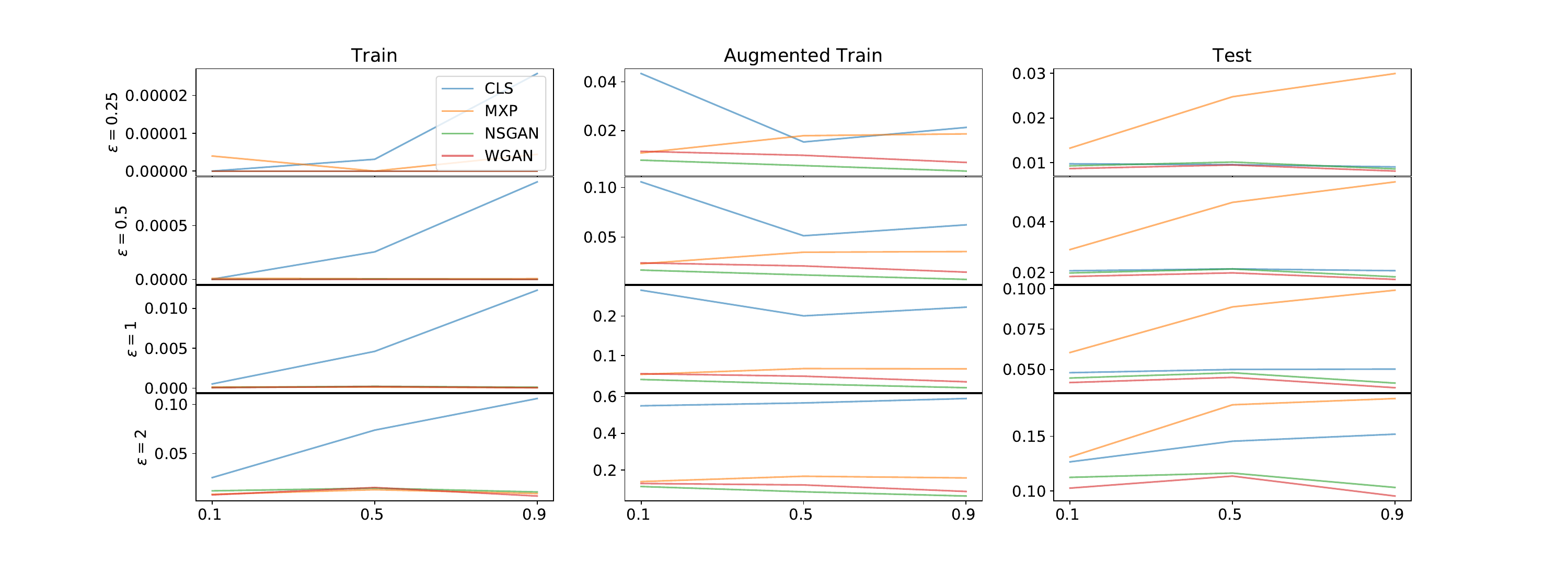}
	\caption{
		Comparison between the average Prediction-Change Stress of models with different augmentations using $l_2$ norm. 
		The y-axis is the  average Stress calculated over all the original training, augmented training, and test set of CIFAR10.
	}
	\label{fig:avg_stress_per_prob}
\end{figure}

Figure~\ref{fig:stress_results} shows the distribution of Prediction-Change Stress (with $\epsilon=1$) over the data points of the original training, augmented training, and test sets. 
In Figure~\ref{fig:avg_stress_per_epsilon} we provide additional plots showing how the prediction-change stress changes when $\epsilon$ is increased.
Also in Figure~\ref{fig:avg_stress_per_prob}, we show how the prediction-change stress changes by increasing the $P_{Aug}$.

We can observe that increasing the augmentation percentage increases the prediction-change stress values around the original training data (sub-figures a,d and g in Figure~\ref{fig:stress_results}).
On the augmented data, sub-figures b,e,h in Figure~\ref{fig:stress_results} show that stress around the augmented data with classical augmentation does not change with increasing the augmentation probability.
The stress increases around Mixup augmented data points and decreases around the data points augmented with GANs. 
Although classical data augmentations results in relatively larger stress around both the original and the augmented training data points, this effect does not translate to the unseen test set.
Mixup augmentation results in the highest prediction-change stress around the testing points as shown in subfigures f,i in Figure~\ref{fig:stress_results}.

\subsubsection{Results with $l_{inf}$ norm}

\begin{figure}[hbt!]
	\centering
	\includegraphics[width=1.03\textwidth]{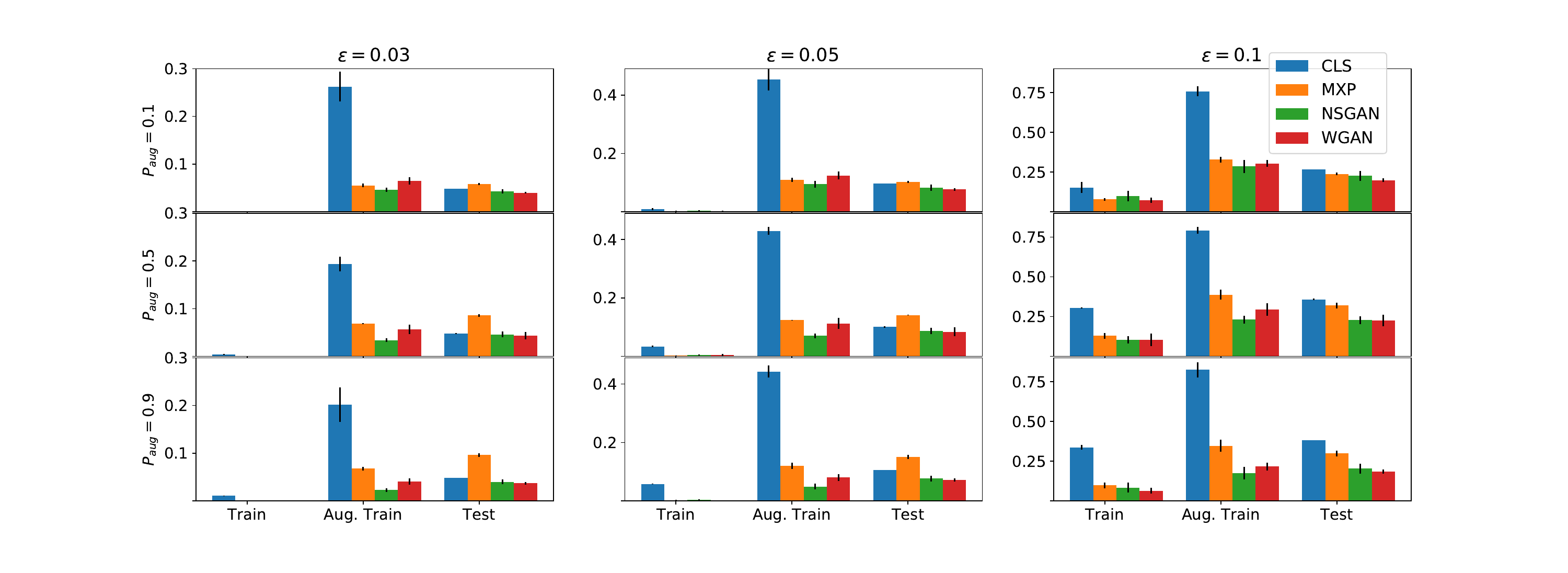}
	\caption{
		Comparison between the Prediction-Change Stress of models with different augmentations using $l_{inf}$ norm.
		The y-axis indicates the different sets.
		The non-Aug and Aug. refer to the training set.
	}
\label{fig:linf_avg_stress_results}
\end{figure}

The prediction-change stress using $\partial B_{\varepsilon} = \{x\in \mathbb{R}^d\mid \max_{k=1}^d x_k = \varepsilon\}$ ($l_{inf}$ norm) are provided in Figure~\ref{fig:linf_avg_stress_results}.
We provide the prediction-change stress for the 50k non-augmented, and 50k augmented examples randomly chosen from the training set, and for all the 10k test examples.
We randomly sample 1k points from the sphere surface $\partial B_{\varepsilon}(\x)$ around each data point $\x$.
We evaluate models trained with all augmentation methods, for $P_{Aug}\in \{0.1, 0.5, 0.9\}$.

\begin{figure}[hbt!]
	\centering
	\includegraphics[width=1.03\textwidth]{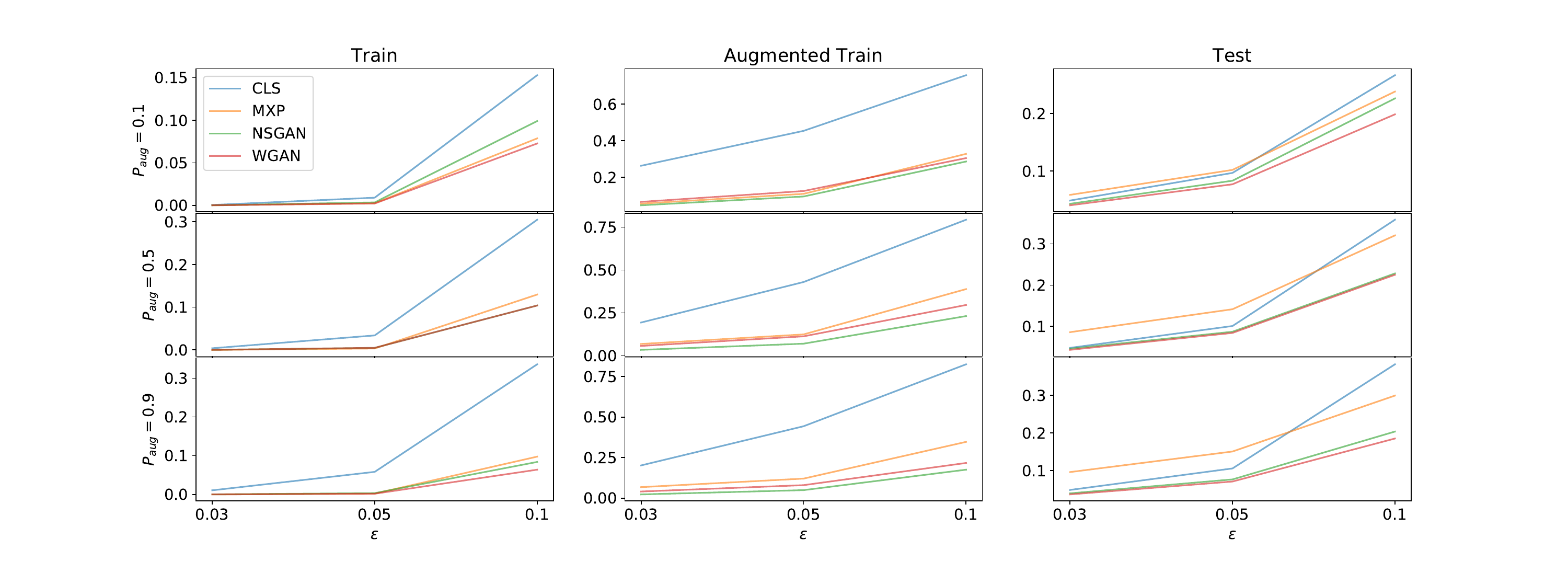}
	\caption{
		Comparison between the Prediction-Change Stress of models with different augmentations using $l_{inf}$ norm. 
		The y-axis is the  average Stress calculated over all the original training, augmented training, and test set of CIFAR10.
	}
	\label{fig:linf_avg_stress_per_epsilon}
\end{figure}

\begin{figure}[hbt!]
	\centering
	\includegraphics[width=1.03\textwidth]{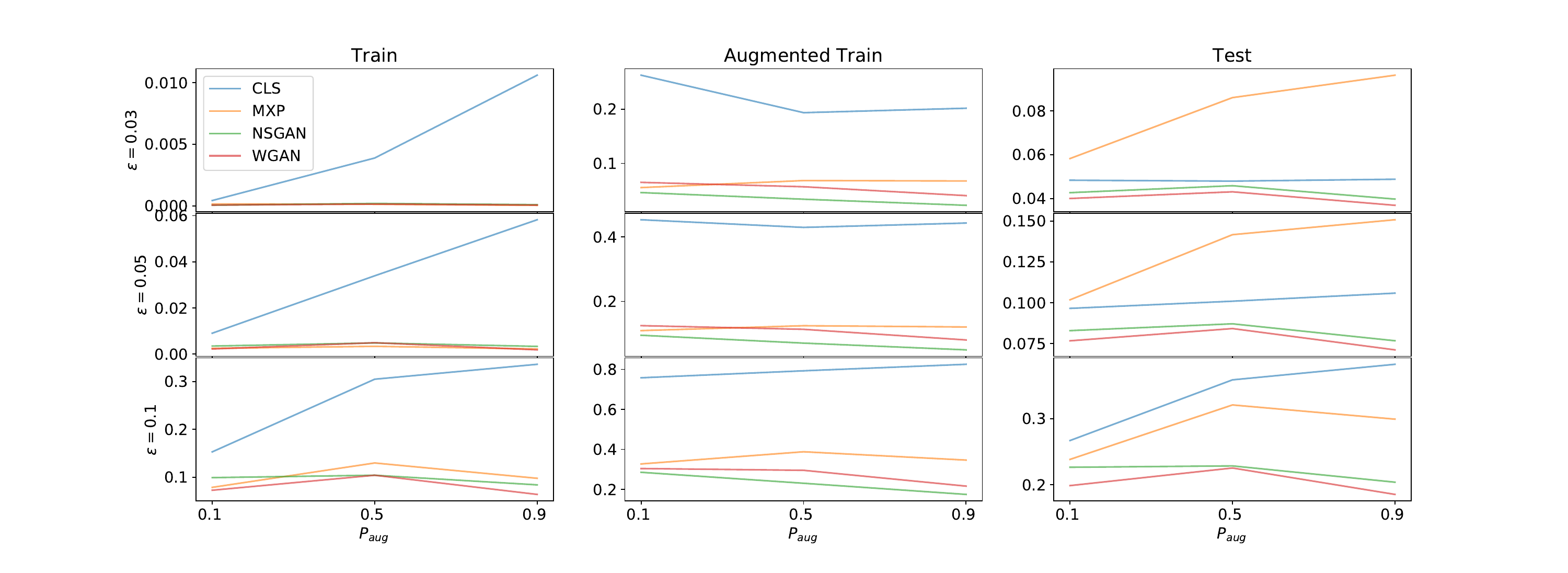}
	\caption{
		Comparison between the average Prediction-Change Stress of models with different augmentations using $l_{inf}$ norm. The y-axis is the  average Stress calculated over all the original training, augmented training, and test set of CIFAR10.
	}
	\label{fig:linf_avg_stress_per_prob}
\end{figure}

\begin{figure}[hbt!]
\centering
    \begin{subfigure}[b]{.3\columnwidth}
           \centering
\includegraphics[width=1.\textwidth]{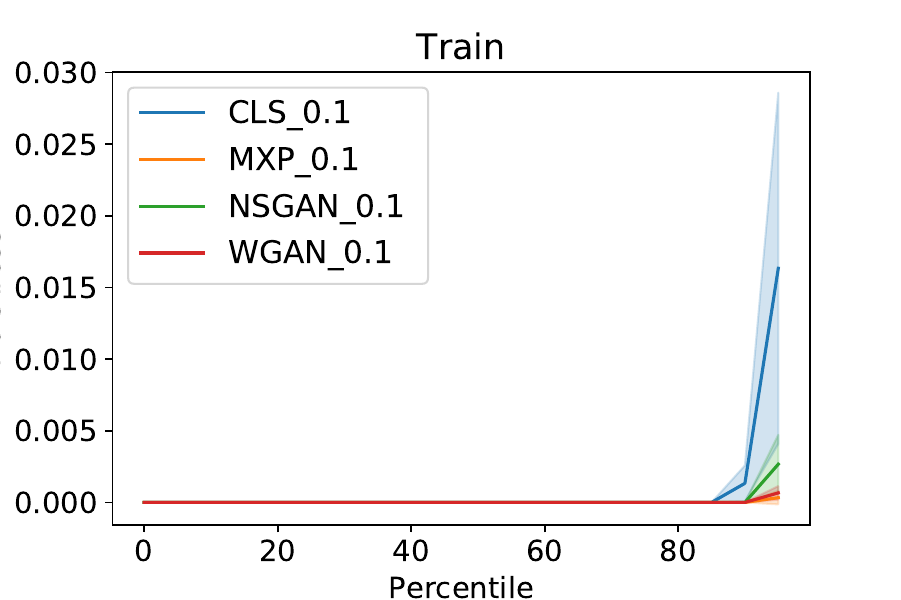}
            \caption{.}
            \label{subfig:}
    \end{subfigure}
    \hfil
    \begin{subfigure}[b]{.3\columnwidth}
           \centering
           \includegraphics[width=1.\textwidth]{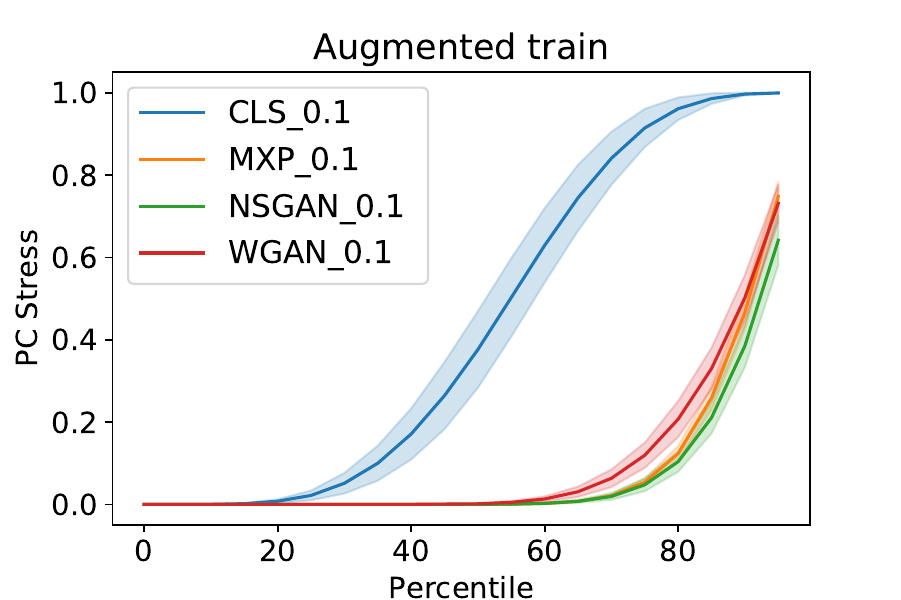}
            \caption{.}
            \label{subfig:}
    \end{subfigure}
    \hfil
    \begin{subfigure}[b]{.3\columnwidth}
           \centering
           \includegraphics[width=1.\textwidth]{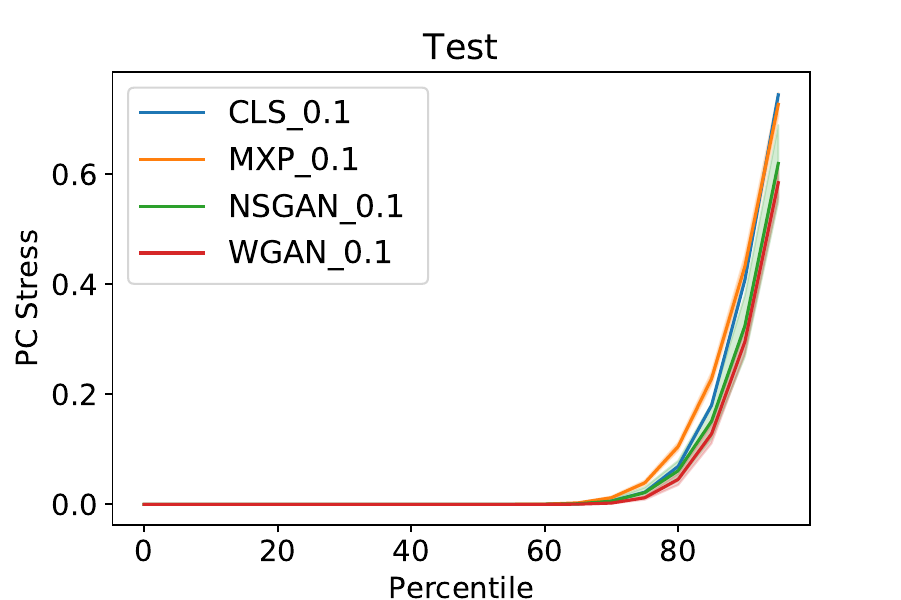}
            \caption{.}
            \label{subfig:}
    \end{subfigure}

    \vspace{1mm}
    \begin{subfigure}[b]{.3\columnwidth}
           \centering
           \includegraphics[width=1.\textwidth]{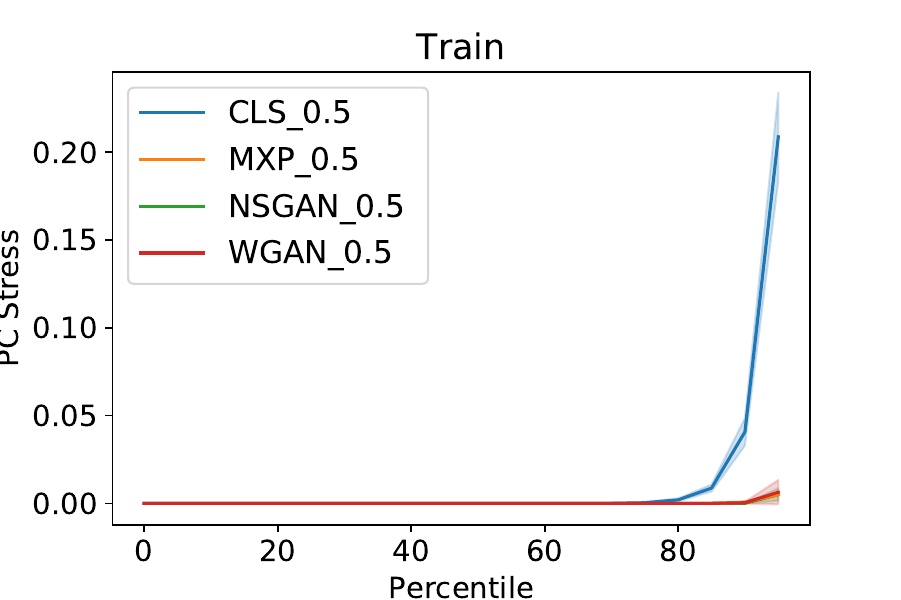}
            \caption{.}
            \label{subfig:}
    \end{subfigure}
    \hfil
    \begin{subfigure}[b]{.3\columnwidth}
           \centering
           \includegraphics[width=1.\textwidth]{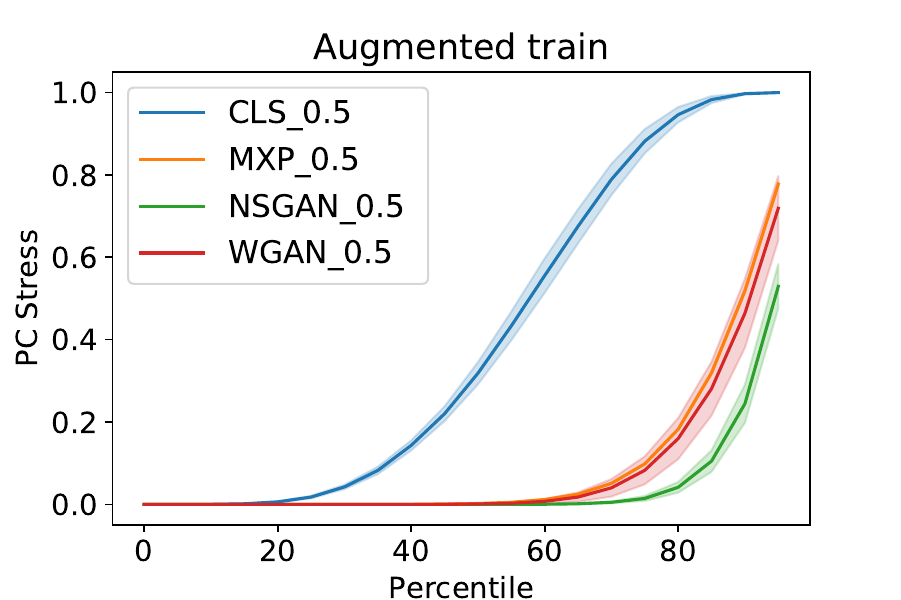}
            \caption{.}
            \label{subfig:}
    \end{subfigure}
  \hfil
    \begin{subfigure}[b]{.3\columnwidth}
           \centering
           \includegraphics[width=1.\textwidth]{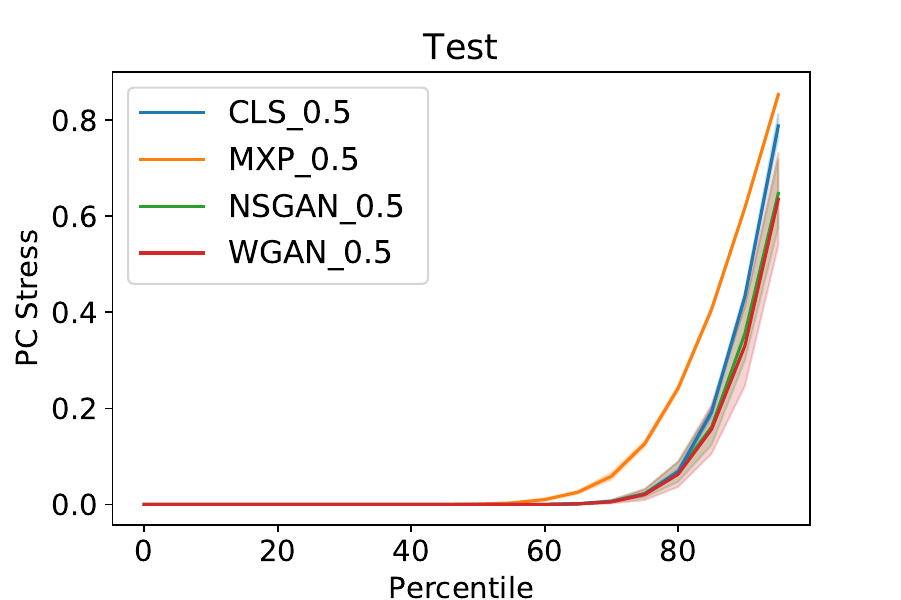}
            \caption{.}
            \label{subfig:}
    \end{subfigure}

    \vspace{1mm}
    \begin{subfigure}[b]{.3\columnwidth}
           \centering
           \includegraphics[width=1.\textwidth]{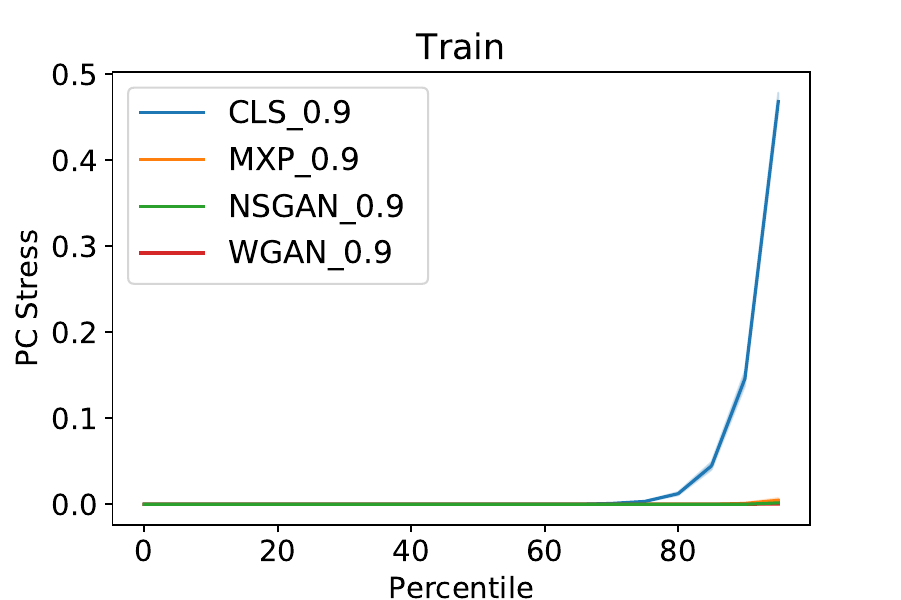}
            \caption{.}
            \label{subfig:}
    \end{subfigure}
    \hfil
    \begin{subfigure}[b]{.3\columnwidth}
           \centering
           \includegraphics[width=1.\textwidth]{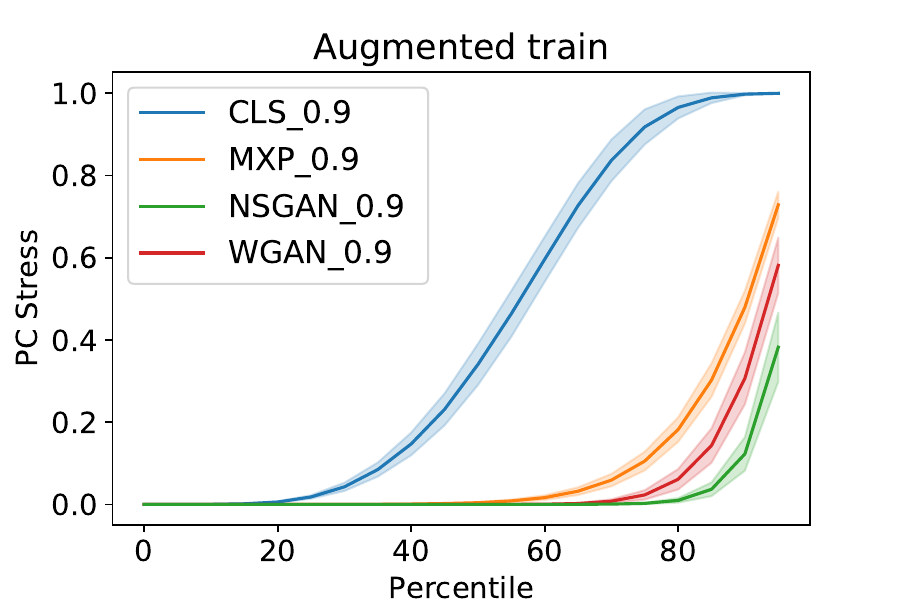}
            \caption{.}
            \label{subfig:}
    \end{subfigure}
    \hfil
    \begin{subfigure}[b]{.3\columnwidth}
           \centering
           \includegraphics[width=1.\textwidth]{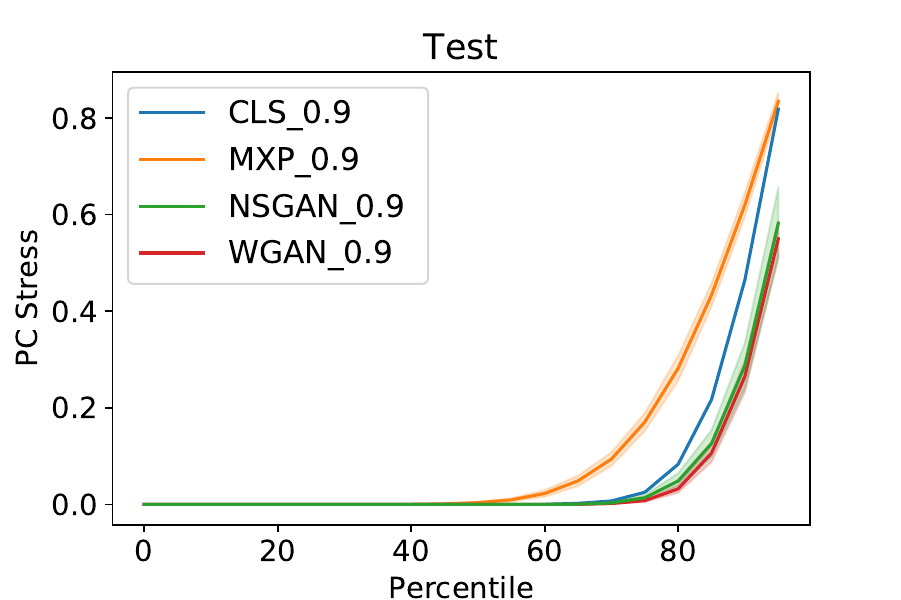}
            \caption{.}
            \label{subfig:}
    \end{subfigure}%

\caption{
Comparison between the distribution Prediction-Change Stress of models with different augmentations using $l_{inf}$ norm, around the original training, the augmented training, and the test set of CIFAR10, and for $\epsilon=0.05$.
First row: 0.1 augmentation.
Second row: 0.5 augmentation.
Third row: 0.9 augmentation.
First column: PCS around the original training set.
Second column: PCS around the augmented training set.
Third column: PCS around the test set.
}
\label{fig:linf_stress_results}
\end{figure}

In Figure~\ref{fig:linf_avg_stress_per_epsilon}, we provide additional plots showing how the prediction-change stress changes when larger $\epsilon$ values are used.
Additionally, we show in Figure~\ref{fig:linf_avg_stress_per_prob}, how the prediction-change stress changes when $P_{Aug}$ is increased.
Finally, in Figure~\ref{fig:linf_stress_results} the plots for prediction-change stress with $\epsilon=0.05$ are provided for different percentiles of the prediction-change stress.
These results outline the distribution of the prediction-change stress in different models.

As can be seen, prediction-change stress with $l_{inf}$ follows a similar trend to the results with the $l_2$ norm  as reported in Section~\ref{subsec:pcs}, as well as the results discussed in Section~\ref{appx:subsec:pce_l2} of the Appendix.

The provided results using various norms, and across different experiments, suggest a strong correlation between the prediction-change stress and risk under adversarial attack, which provides more evidence for the prediction-change stress to be used as a new measure for evaluating models in the adversarial setting.

\newpage\clearpage

\section{Network Architectures}
\label{appx:network_architectures}
The network architectures used in the GAN models are provided in Table~\ref{appx:tab:gan_architectures}.

\begin{table}[ht]
\caption{The architectures of the generator (left) and the discriminator (right).}
\label{appx:tab:gan_architectures}
\centering
\resizebox{.4\columnwidth}{!}{
\begin{tabular}{|l|c|c|}
\hline
Gen. Layer (type)&Shape&Param \# \\ \hline\hline
            Linear-1&                 [1024]   &      113,664\\
       BatchNorm2d-2&             [ 64, 4, 4]   &          128\\
              ReLU-3&             [ 64, 4, 4]   &            0\\
          Upsample-4&             [ 64, 8, 8]   &            0\\
            Conv2d-5&             [ 64, 8, 8]   &       36,928\\
       BatchNorm2d-6&             [ 64, 8, 8]   &          128\\
              ReLU-7&             [ 64, 8, 8]   &            0\\
            Conv2d-8&             [ 64, 8, 8]   &       36,928\\
          Upsample-9&             [ 64, 8, 8]   &            0\\
ResBlockGenerator-10&             [ 64, 8, 8]   &            0\\
      BatchNorm2d-11&             [ 64, 8, 8]   &          128\\
             ReLU-12&             [ 64, 8, 8]   &            0\\
         Upsample-13&           [ 64, 16, 16]   &            0\\
           Conv2d-14&           [ 64, 16, 16]   &       36,928\\
      BatchNorm2d-15&           [ 64, 16, 16]   &          128\\
             ReLU-16&           [ 64, 16, 16]   &            0\\
           Conv2d-17&           [ 64, 16, 16]   &       36,928\\
         Upsample-18&           [ 64, 16, 16]   &            0\\
ResBlockGenerator-19&           [ 64, 16, 16]   &            0\\
      BatchNorm2d-20&           [ 64, 16, 16]   &          128\\
             ReLU-21&           [ 64, 16, 16]   &            0\\
         Upsample-22&           [ 64, 32, 32]   &            0\\
           Conv2d-23&           [ 64, 32, 32]   &       36,928\\
      BatchNorm2d-24&           [ 64, 32, 32]   &          128\\
             ReLU-25&           [ 64, 32, 32]   &            0\\
           Conv2d-26&           [ 64, 32, 32]   &       36,928\\
         Upsample-27&           [ 64, 32, 32]   &            0\\
ResBlockGenerator-28&           [ 64, 32, 32]   &            0\\
      BatchNorm2d-29&           [ 64, 32, 32]   &          128\\
             ReLU-30&           [ 64, 32, 32]   &            0\\
           Conv2d-31&            [ 3, 32, 32]   &        1,731\\
             Tanh-32&            [ 3, 32, 32]   &            0\\
\hline

\end{tabular}}
\quad
\resizebox{.4\columnwidth}{!}{
\begin{tabular}{|l|c|c|}
\hline
Disc. Layer (type)&Shape&Param \# \\ \hline\hline
       BatchNorm2d-1          &  [ 3, 32, 32]       &        6\\
            Conv2d-2          & [ 64, 32, 32]       &    1,792\\
            Conv2d-3          & [ 64, 32, 32]       &    1,792\\
              ReLU-4          & [ 64, 32, 32]       &        0\\
            Conv2d-5          & [ 64, 32, 32]       &   36,928\\
            Conv2d-6          & [ 64, 32, 32]       &   36,928\\
         AvgPool2d-7          & [ 64, 16, 16]       &        0\\
         AvgPool2d-8          &  [ 3, 16, 16]       &        0\\
            Conv2d-9          & [ 64, 16, 16]       &      256\\
           Conv2d-10          & [ 64, 16, 16]       &      256\\
FirstResBlockDiscriminator-11 &         [ 64, 16, 1&6]      0\\
             ReLU-12          & [ 64, 16, 16]       &        0\\
           Conv2d-13          & [ 64, 16, 16]       &   36,928\\
           Conv2d-14          & [ 64, 16, 16]       &   36,928\\
             ReLU-15          & [ 64, 16, 16]       &        0\\
           Conv2d-16          & [ 64, 16, 16]       &   36,928\\
           Conv2d-17          & [ 64, 16, 16]       &   36,928\\
        AvgPool2d-18          &   [ 64, 8, 8]       &        0\\
           Conv2d-19          & [ 64, 16, 16]       &    4,160\\
           Conv2d-20          & [ 64, 16, 16]       &    4,160\\
        AvgPool2d-21          &   [ 64, 8, 8]       &        0\\
ResBlockDiscriminator-22      &       [ 64, 8, 8]   &        0\\
             ReLU-23          &   [ 64, 8, 8]       &        0\\
           Conv2d-24          &   [ 64, 8, 8]       &   36,928\\
           Conv2d-25          &   [ 64, 8, 8]       &   36,928\\
             ReLU-26          &   [ 64, 8, 8]       &        0\\
           Conv2d-27          &   [ 64, 8, 8]       &   36,928\\
           Conv2d-28          &   [ 64, 8, 8]       &   36,928\\
        AvgPool2d-29          &   [ 64, 4, 4]       &        0\\
           Conv2d-30          &   [ 64, 8, 8]       &    4,160\\
           Conv2d-31          &   [ 64, 8, 8]       &    4,160\\
        AvgPool2d-32          &   [ 64, 4, 4]       &        0\\
ResBlockDiscriminator-33      &      [ 64, 4, 4]   &        0\\
             ReLU-34          &   [ 64, 4, 4]       &        0\\
           Conv2d-35          &   [ 64, 4, 4]       &   36,928\\
           Conv2d-36          &   [ 64, 4, 4]       &   36,928\\
             ReLU-37          &   [ 64, 4, 4]       &        0\\
           Conv2d-38          &   [ 64, 4, 4]       &   36,928\\
           Conv2d-39          &   [ 64, 4, 4]       &   36,928\\
ResBlockDiscriminator-40      &      [ 64, 4, 4]   &        0\\
             ReLU-41          &   [ 64, 4, 4]       &        0\\
           Conv2d-42          &   [ 64, 4, 4]       &   36,928\\
           Conv2d-43          &   [ 64, 4, 4]       &   36,928\\
             ReLU-44          &   [ 64, 4, 4]       &        0\\
           Conv2d-45          &   [ 64, 4, 4]       &   36,928\\
           Conv2d-46          &   [ 64, 4, 4]       &   36,928\\
ResBlockDiscriminator-47      &       [ 64, 4, 4]   &        0\\
             ReLU-48          &   [ 64, 4, 4]       &        0\\
        AvgPool2d-49          &   [ 64, 1, 1]       &        0\\
           Linear-50          &          [ 1]       &       65\\
ResBlockClassifier-47      &       [ 64, 4, 4]   &        0\\
             ReLU-48          &   [ 64, 4, 4]       &        0\\
        AvgPool2d-49          &   [ 64, 1, 1]       &        0\\
           Linear-50          &         [ 10]       &      650\\
\hline
\end{tabular}
}
\end{table}

\section{Tools and Libraries}
\label{appx:sec:tools_libs}
All experiments have been implemented in python using \verb!pytorch lib!~\cite{paszke2019pytorch}.
The adversarial attacks are done using the \verb!robustness lib!~\cite{robustness}.
For logging the experiments and controlling hyperparameters and randomness, \verb!sacred lib!~\cite{greff2017sacred} is used.
For analysing the results, \verb!incense lib!~\cite{incense} was used to fetch the results from the \verb!sacred lib! database.

\end{document}